\documentclass[11pt]{article}

\usepackage[preprint]{acl}

\usepackage{times}
\usepackage{latexsym}
\usepackage{booktabs}
\usepackage{tikz}
\usetikzlibrary{positioning,arrows.meta,fit,backgrounds}
\usepackage{pgfplots}

\usepackage[T1]{fontenc}

\usepackage[utf8]{inputenc}

\usepackage{microtype}

\usepackage{inconsolata}

\usepackage{graphicx}
\usepackage{amsmath}
\usepackage{enumitem}
\pgfplotsset{compat=1.18}

\title{When Does Consistency Pay Off? Relational Constraints for LLM-based Document-Level Relation Extraction under Incomplete Annotations}

\author{Mingxuan Sun}

\begin{document}
\maketitle
\begin{abstract}
Consistency-based repair is widely assumed to improve LLM-based document-level relation extraction (DocRE), but this assumption can fail when the benchmark itself is incompletely annotated.
Whether such repair helps depends strongly on how closed the gold annotations are under the constraints it enforces.
We introduce $\Delta_{\mathrm{closure}}$, the rate at which gold annotations lack facts implied by the very rules used for repair, computable from the annotations and rules alone before any repair is run.
The same typed constraints drive \textsc{Consistre}, which applies them at inference time (constraint feedback for reflection, deterministic closure) and at training time (distillation and consistency-aware reinforcement learning).
On 998 DocRED documents, whose Re-DocRED revision lowers $\Delta_{\mathrm{closure}}$ from 1.71\% to 0.257\%, rescoring identical predictions reverses the method ranking: Self-Refine leads under DocRED, while a constraint-feedback variant moves from fifth place to first.
On the near-closed DWIE, constraint feedback with chain-of-thought beats three generic reflection baselines by 0.046--0.107 micro-F1 ($p{=}0.0002$), and deterministic closure of the base output scores higher still at 3.6$\times$ lower cost.
A reinforcement-learning gain on DocRED ($+0.025$) does not survive Re-DocRED rescoring.
Consistency repair pays off where annotations are closed, and $\Delta_{\mathrm{closure}}$ tells in advance whether they are.
\end{abstract}

\section{Introduction}
\label{sec:intro}

Document-level relation extraction (DocRE) identifies the relations that hold among the entities of a document~\cite{yao2019docred}. With large language models (LLMs), it is increasingly cast as generating relational triples directly~\cite{wadhwa2023revisiting,wan2023gptre,xue2024autore}. What the model returns, however, is not a set of independent triples but a relational graph, and such graphs have structure: symmetric and inverse relations come in pairs, relations compose and imply one another, and a single-valued attribute takes one value. An LLM decides each entity pair largely in isolation, so nothing in decoding enforces this structure: on DocRED, our GPT-5.2 baseline leaves 90\% of the rule instances triggered by its own output unclosed.

Relational constraints therefore look like an obvious way to repair the graph. Logic rules have long been used to improve supervised DocRE models~\cite{ru-etal-2021-learning,fan-etal-2022-boosting,qi-etal-2024-end}. For LLMs, they enter as rule constraints on extraction~\cite{zhang2025logical}, as consistency feedback for iterative refinement~\cite{wang2026document,han2025cognitive,wang2025leveraging}, or as logic-aware training rewards~\cite{xu2026logicreward}. Generic self-refinement methods, Self-Refine~\cite{madaan2023self}, Reflexion~\cite{shinn2023reflexion} and CoVe~\cite{dhuliawala2024chain}, iterate the same way without an external verification signal. The shared assumption is that a more consistent graph is a more correct one, with the gold annotation as a neutral judge.

But benchmark annotations may themselves violate these constraints. DocRED is known to miss many true relation instances~\cite{huang2022does}: under the rules we apply to model outputs, its development gold leaves 18\% of the rule instances it triggers unclosed, against 3\% for the same documents in Re-DocRED~\cite{tan2022revisiting}. When the gold states $(a,\textit{spouse},b)$ but omits $(b,\textit{spouse},a)$, a repair that restores symmetry is scored as a false positive, and a reflection step that deletes a true but unannotated triple is scored as a correction. We therefore ask: \emph{if the gold violates the same constraints used to repair predictions, when does enforcing consistency help rather than hurt?}

We answer it by treating the constraints as an external instrument that scores predictions and gold alike. We formalize relational consistency as five constraint classes (typed Horn rules for symmetry, inversion, composition and implication, plus an at-most-one functional constraint), checked by one deterministic enumerator and grounded per benchmark: ten Wikidata-derived rules for DocRED~\cite{vrandevcic2014wikidata} and the 25 rules published with DWIE~\cite{zaporojets2021dwie}, each with functional constraints. Applied to the gold, the enumerator yields $\Delta_{\mathrm{closure}}$, the number of rule-implied facts missing per gold triple. Applied to predictions, it drives \textsc{Consistre}, which uses the constraints at inference time (constraint cues, reflection over violation feedback, deterministic closure) and at training time (distillation, then GRPO~\cite{shao2024deepseekmathpushinglimitsmathematical} with a consistency-aware reward). We evaluate GPT-5.2 and Qwen3-8B on 998 DocRED development documents, scoring identical predictions under DocRED and Re-DocRED so that only the gold changes, and on 200 DWIE documents. We find:

\begin{itemize}[leftmargin=*,itemsep=1pt,topsep=2pt,parsep=0pt]
\item \textbf{Annotation closure predicts effectiveness.} Deterministic closure adds facts as precise as the model's own predictions under Re-DocRED and DWIE (ratio 0.99) but only 68--72\% as precise under DocRED, where many are true but unannotated.
\item \textbf{The same predictions reverse their ranking under revised annotation.} A constraint-feedback variant ranks fifth under DocRED ($-0.035$ vs.\ Self-Refine) and first under Re-DocRED ($+0.019$ vs.\ Reflexion), both at $p{=}0.0002$. Self-Refine's loop gain turns from $+0.027$ to $-0.017$ as its deletions go from rewarded to penalized: 14\% are true relations under DocRED, 31\% under Re-DocRED.
\item \textbf{Where the gold is near-closed, deterministic closure is the baseline to beat before adding an LLM reflection loop.} On DWIE, constraint feedback with chain-of-thought beats all three generic baselines by at least 0.046 ($p{=}0.0002$), yet closing the base output under the rules scores higher still (0.710 vs.\ 0.692 over three seeds) at 3.6$\times$ lower cost; under Re-DocRED, the generic loops fall below even the single-pass base output, at 3.6--4.2 times its cost per document (Figure~\ref{fig:cost}).
\item \textbf{Training-time gains can partly exploit annotation gaps.} GRPO adds $+0.025$ on DocRED ($p{=}0.0002$) but $-0.005$ ($p{=}0.16$) under Re-DocRED, and about 20\% of the predictions it prunes are true relations missing from DocRED's gold. On DWIE, its predictions shrink by only 1.7\% and precision and recall rise together.
\end{itemize}
Consistency repair is thus conditional, and $\Delta_{\mathrm{closure}}$ measures its condition before any system is run.

\section{Related Work}
\label{sec:related}

\paragraph{DocRE, logic rules, and annotation gaps.}
Supervised DocRE models reason over mention graphs~\cite{zeng2020gain}, adaptive thresholds~\cite{zhou2021atlop}, evidence~\cite{ma2023dreeam}, or distilled distant supervision~\cite{tan-etal-2022-document}; Appendix~\ref{app:sota} compares them with our setting. Rule-based variants learn or mine logic rules from the training annotations---LogiRE~\cite{ru-etal-2021-learning}, MILR~\cite{fan-etal-2022-boosting}, BCBR~\cite{liu-etal-2023-document-level}, JMRL~\cite{qi-etal-2024-end}---and so inherit whatever gaps the gold contains; the gold itself is never measured. Those gaps are documented: DocRED misses many true relation instances~\cite{huang2022does}, Re-DocRED revises its annotations~\cite{tan2022revisiting}, Re$^2$-DocRED expands them further with an LLM pipeline over an enriched rule schema~\cite{heng-etal-2026-re2}, and DWIE~\cite{zaporojets2021dwie} offers an entity-centric alternative with near-exhaustive annotation.

\paragraph{LLM extraction and refinement.}
LLMs recast DocRE as generation~\cite{wadhwa2023revisiting,xu2024large,wan2023gptre,xue2024autore}, and consistency signals have been used to retrieve and denoise knowledge~\cite{sun2024consistency}. Generic self-refinement~\cite{madaan2023self,shinn2023reflexion,dhuliawala2024chain} can degrade without external verification~\cite{huang2024large}. DocRE-specific frameworks add consistency signals: BA-MCTS~\cite{wang2026document} searches add/delete/modify actions under an LLM-assigned score, ToolRE~\cite{wang2025leveraging} verifies triples with LLM-generated scores, and Cognitive Mirroring~\cite{han2025cognitive} iterates with rule-based and reconstruction feedback. Our check is instead deterministic rule matching with full coverage, and its feedback names the rule that fails and the fact that is missing.

\paragraph{Logic constraints for LLMs.}
Logic has been injected into neural models at least since \citet{li2019augmenting}. For LLMs, constraints serve as a \emph{measure} of a model's own consistency (REPAIR;~\citealp{liu2024aligning}), as a \emph{loss} (LOCO-LMs;~\citealp{calanzone2025logically}), as \emph{injected knowledge} (RuAG;~\citealp{zhang2025ruag}), or as a \emph{reward} (LogicReward;~\citealp{xu2026logicreward}). All of these measure or improve the model. We point the rules at the gold annotations instead, and treat them as reproducible instruments rather than truths that annotations should be expanded to satisfy (Appendix~\ref{app:related}).

\begin{figure*}[t]
\centering
\includegraphics[width=0.95\textwidth]{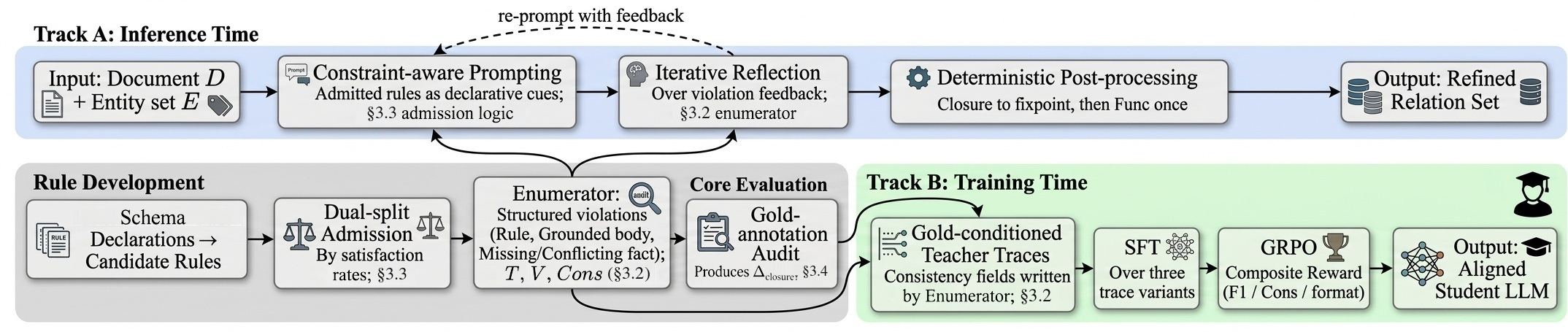}
\caption{Overview of \textsc{Consistre}. Rule Development admits the constraint set; the Enumerator checks any triple set against it, audits the gold ($\Delta_{\mathrm{closure}}$), and drives Track~A (inference time) and Track~B (training time). Section~\ref{sec:formal} introduces each module.}
\label{fig:overview}
\end{figure*}

\section{Relational Consistency: Formalization and Measurement}
\label{sec:formal}
\label{sec:formal:task}

\paragraph{Problem definition.}
Given a document $D$ with entity clusters $\mathcal{E}$ and a relation inventory $\mathcal{R}$, a DocRE system $f$ returns
\begin{equation}
\hat{\mathcal{T}} = f(D,\mathcal{E}) \subseteq \mathcal{E}\times\mathcal{R}\times\mathcal{E},
\label{eq:task}
\end{equation}
scored by micro-F1 against the gold triples $\mathcal{G}$~\cite{yao2019docred}. A constraint set $\mathcal{C}=\mathcal{C}^{+}\cup\mathcal{C}^{-}$ of positive rules and functional constraints states which triples must co-occur and which must not. We ask when repairing $\hat{\mathcal{T}}$ toward $\mathcal{C}$ raises F1 against $\mathcal{G}$, and whether auditing $\mathcal{G}$ against $\mathcal{C}$ predicts it.

\paragraph{Overview.}
In Figure~\ref{fig:overview}, \emph{Rule Development} (bottom left) admits $\mathcal{C}$ for each benchmark (Section~\ref{sec:formal:grounding}); the \emph{Enumerator} (center) returns the triggered and violated instantiations of any triple set, which score predictions as Cons (Section~\ref{sec:formal:cons}) and the gold as $\Delta_{\mathrm{closure}}$ in the \emph{Gold-annotation Audit} (Section~\ref{sec:formal:audit}); and \textsc{Consistre} uses the same enumerator at inference time (Track~A, top) and at training time (Track~B, bottom right; Section~\ref{sec:framework}).

\subsection{Five Constraint Classes}
\label{sec:formal:rules}

\emph{Positive rules} are Horn rules over binary relations with one or two body atoms, in four classes:
\begin{equation}
\begin{aligned}
\text{Sym:}\quad & r(x,y)\Rightarrow r(y,x),\\
\text{Inv:}\quad & r_1(x,y)\Rightarrow r_2(y,x),\quad r_1\neq r_2,\\
\text{Comp:}\quad & r_1(x,y)\wedge r_2(y,z)\Rightarrow r_3(x,z),\\
\text{Imp:}\quad & r_1(x,y)\Rightarrow r_2(x,y),
\end{aligned}
\label{eq:pos}
\end{equation}
with transitivity as the special case $r_1{=}r_2{=}r_3$ of Comp. Examples are \textit{spouse} (Sym), \textit{father}$\Rightarrow$\textit{child} (Inv), \textit{capital of}$\Rightarrow$\textit{located in} (Imp), and DWIE's \texttt{in2}$\,\circ\,$\texttt{in0}$\Rightarrow$\texttt{in0} (Comp; Section~\ref{sec:formal:grounding}). A rule may carry an entity-type guard and is triggered only where the guard holds; on DWIE, \texttt{member\_of}$\Rightarrow$\texttt{player\_of} applies only when the subject is typed \texttt{sport\_player} (Appendix~\ref{app:rules}). The \emph{negative} class is the functional constraint $\mathrm{Func}(r)$:
\begin{equation}
r(x,y_1)\wedge r(x,y_2)\wedge y_1\neq y_2\Rightarrow\bot,
\label{eq:func}
\end{equation}
i.e., at most one tail per \emph{anchor} $(r,x)$, as in DWIE's \texttt{in0}: a place lies in one country.

\subsection{Violation Enumeration and Cons}
\label{sec:formal:cons}

A deterministic enumerator grounds every constraint $c$ over a triple set $\mathcal{T}$ and returns its triggered instantiations $T_c(\mathcal{T})$ and violations $V_c(\mathcal{T})$: for a positive rule, the instantiations whose body and guard hold and those among them whose head is missing; for $\mathrm{Func}(r)$, whose every body instantiation is already a violation, the anchors $(r,x)$ with at least one tail and those with two or more. Pooling the counts over constraints and documents gives
\begin{equation}
\mathrm{Cons}(\mathcal{T}) = 1-\frac{\sum_{c}|V_c(\mathcal{T})|}{\sum_{c}|T_c(\mathcal{T})|}.
\label{eq:cons}
\end{equation}

Positive and negative constraints never share a denominator, so we report $\mathrm{Cons}_{\mathrm{pos}}$ and $\mathrm{Cons}_{\mathrm{func}}$ separately. A document that triggers no rule has undefined Cons instead of 1.0, and we also report the zero-trigger rate. For training reward calculations, the Cons term is neutralized for such outputs. This is because scoring these outputs as either 1 or 0 would encourage the policy to make fewer or more predictions (Appendix~\ref{app:eval}). Both tracks use this same enumerator and normalization without modification.

\subsection{Per-Benchmark Grounding}
\label{sec:formal:grounding}

Positive rules originate from external declarations, and functional constraints are derived from single-value semantics. Table~\ref{tab:provenance} tracks the source of each rule set. For DocRED, we generate 103 candidates covering all 96 relation types using Wikidata property declarations~\cite{vrandevcic2014wikidata} (snapshot 2026-08-26). We accept a rule only when it has at least 20 triggered instantiations on the dev set and a satisfaction rate of no less than 0.50 on \emph{both} the dev and train sets. Ten rules pass (2 Sym, 7 Inv, 1 Imp, 0 Comp); functional constraints come from Wikidata's single-value declarations, tiered by their wording (3 hard, 15 soft). For DWIE, the dataset authors' rule appendix~\cite{zaporojets2021dwie} yields 25 rules (2 Sym, 3 Inv, 13 Imp, 7 Comp) after removing its 12 nominal-variant (\texttt{-x}) relation types. DWIE declares no functional constraint, so we wrote down one criterion---within a document, the head entity determines the tail---and applied it to 23 candidate relations, admitting 10 and recording a reason for each of the 13 exclusions (e.g., dual citizenship rules out \texttt{citizen\_of}; Appendix~\ref{app:dwiefunc}). Their direct effect on the results is negligible: they license no deletion in any reflection loop and remove at most one prediction per post-processed GPT-5.2 run on DWIE; we did not ablate them as prompt cues. All rule sets were finalized before any of the experiments reported here. No rule was chosen based on model predictions or model performance. DWIE’s 1,119 implicit entities, which are linked but never explicitly mentioned, are retained in full. No benchmark covers all five classes. Therefore, grounding must be re-run for each benchmark.

\subsection{Gold Audit and \texorpdfstring{$\Delta_{\mathrm{closure}}$}{Delta-closure}}
\label{sec:formal:audit}

Because the rules are reproducible, they can be pointed at the gold annotations. We define the closure-defect rate
\begin{equation}
\Delta_{\mathrm{closure}} = |M(\mathcal{G})|\,/\,|\mathcal{G}|,
\label{eq:delta}
\end{equation}
where $M(\mathcal{G})$ is the set of distinct head facts missing from $\mathcal{G}$, i.e., the heads of $\bigcup_{c\in\mathcal{C}^{+}}V_c(\mathcal{G})$, and report it for both annotation versions without asserting which is correct. We state the apparent circularity upfront: admission selects rules that the annotations largely satisfy, and $\Delta_{\mathrm{closure}}$ measures how far they still fall short. It should be read as ``on rules the annotation practice demonstrably follows, the gold lacks $N$ implied facts per 100 triples''---a statement about annotation closure, not about rule truth.

Table~\ref{tab:audit} gives the audits. Absolute values are not comparable across benchmarks, whose rule sets differ; the decisive comparison is paired. On the same 998 documents, Re-DocRED's revision~\cite{tan2022revisiting} shrinks $\Delta_{\mathrm{closure}}$ from 1.71\% to 0.257\% ($6.7\times$), while functional conflicts move the other way (soft-Func Cons $0.9870\rightarrow0.9743$; net violations $17\rightarrow61$): revision closes positive-rule gaps but adds single-value conflicts, a divergence visible only because the two families are scored separately. The revision is not itself rule-driven: of the 22{,}754 triples Re-DocRED adds on these documents, only 6.8\% are derivable by our ten rules, almost all of them the reverse direction of a symmetric or inverse pair, and the largest block of additions (\texttt{P131}, 29\%) is only 10.8\% derivable (Appendix~\ref{app:a11}). Per-rule audits are in Appendix~\ref{app:rules}.

\begin{table}[t]
\caption{Gold audit. $\Delta_{\mathrm{closure}}$: distinct missing head facts per gold triple. Positive and negative constraints are scored separately; values are not comparable across benchmarks, whose rule sets differ. The first two rows are the same documents under two annotation versions.}
\label{tab:audit}
\centering\small
\resizebox{\columnwidth}{!}{%
\begin{tabular}{lcccc}
\toprule
Corpus & $\Delta_{\mathrm{closure}}$ & Cons$_{\mathrm{pos}}$ & Cons$_{\mathrm{func}}$ & net viol. \\
\midrule
DocRED (dev 998)     & 1.71\% & 0.8175 & 0.9870 (soft) & 17 \\
Re-DocRED (same 998) & 0.257\% & 0.9681 & 0.9743 (soft) & 61 \\
DWIE (802)           & 0.55\% & 0.9764 & 0.9992 & 4 \\
\bottomrule
\end{tabular}}
\end{table}

\section{\textsc{Consistre}: Two Instantiations of One Formulation}
\label{sec:framework}
\label{sec:framework:overview}

\textsc{Consistre} instantiates Section~\ref{sec:formal} in two tracks that share its rules, enumerator, and Cons definitions, so both report Cons values computed by the same code.

\subsection{Track A: Inference-Time Consistency}
\label{sec:framework:tracka}

Track~A (Figure~\ref{fig:overview}, top) works with black-box LLMs in four steps. \emph{Constraint-aware prompting}: the admitted rules are rendered as declarative cues in the extraction prompt (e.g., ``if you extract \emph{spouse}$(x,y)$, also check \emph{spouse}$(y,x)$''); they describe the constraint surface only, reveal no gold content, and never prescribe deletions (Appendix~\ref{app:prompts}). \emph{Verification}: the enumerator checks the output and returns structured feedback---for each violated instantiation, the rule, its grounded body, and the missing head, or for Func the conflicting pair---rather than a scalar score. \emph{Iterative reflection}: the model revises its output against this feedback until no violation remains or three rounds have run. \emph{Deterministic post-processing}: after the final round only,
\begin{equation}
\hat{\mathcal{T}}_{\mathrm{post}} = \mathrm{Func}\big(\mathrm{Cl}_{\mathcal{C}^{+}}(\hat{\mathcal{T}})\big),
\label{eq:post}
\end{equation}
where $\mathrm{Cl}_{\mathcal{C}^{+}}$ adds positive-rule heads until a fixpoint and $\mathrm{Func}$ then applies the functional constraints once. Closure runs first because Horn rules cannot introduce contradictions, so Func adjudicates any collision with closure-added triples; on DocRED, only the three hard functional constraints are enforced (Appendix~\ref{app:postproc}).

The configurations combine these steps. \textsc{base} asks only for the triples; \textsc{cot} adds a reasoning field; \textsc{sr} adds the cues, a \texttt{consistency\_check} field, and reflection with enumerator feedback; \textsc{cot+sr} combines the two; \textsc{cues-only} adds the cues alone, without reflection or post-processing.

\subsection{Track B: Training-Time Consistency}
\label{sec:framework:trackb}

Track~B (Figure~\ref{fig:overview}, bottom right) distills a gold-conditioned teacher into relatively small student LLMs and then refines them with reinforcement learning, using group relative policy optimization (GRPO)~\cite{shao2024deepseekmathpushinglimitsmathematical}. For the \emph{teacher traces}, the teacher sees each document \emph{with} its gold triples, explains their evidence, and echoes the relation field verbatim (99.9\% exact). Its \texttt{consistency\_check} field is populated using the enumerator's output instead of the teacher’s independent judgment. Without this support, the teacher identified only 1 out of the 10 violations the enumerator found in a 20-document probe. When provided with the enumerator’s output, however, the teacher recorded 98\% of all enumerated violations across 1{,}200 documents (386/392). Gold triples remain unchanged, even when the check determines that the gold data itself breaks a rule. In \emph{SFT}, following rationale distillation~\cite{hsieh2023distillingstepbystepoutperforminglarger}, students are fine-tuned with LoRA~\cite{hu2021lora} on three variants of the same documents: \emph{full} traces (reasoning, check, relations), \emph{no-cons-field} (reasoning and relations), and \emph{gold-only} (relations only). \emph{GRPO} then continues from a fine-tuned student without a learned value model: for each training document, it samples a group of completions, scores each with
\begin{equation}
\begin{split}
R(y) ={}& \lambda_F\,\mathrm{F1}(y) + \lambda_C\,\mathrm{Cons}(y)\\
&+ \lambda_M\,\mathrm{Fmt}(y),
\end{split}
\label{eq:reward}
\end{equation}
and updates the student with a clipped policy-gradient objective in which each completion's advantage is its reward relative to the mean of its group. F1 is computed against the document's gold triples, Fmt scores schema-conformant output, and the Cons term is neutralized on completions that trigger no rule (weights in Section~\ref{sec:setup:systems}); we report learning rates $1{\times}10^{-5}$ and $5{\times}10^{-5}$. Training details are in Appendix~\ref{app:training}.

\section{Experimental Setup}
\label{sec:setup}

\begin{table*}[t]
\caption{Track A on DocRED dev (998 documents, post-processing off); F1 (Re-DocRED) rescores the identical predictions, and the best F1 under each gold is in bold. Cons is measured before post-processing, with triggered/violated instantiations $T$/$V$ and the zero-trigger rate (ZT). Ign F1 drops gold triples seen in training (Appendix~\ref{app:eval}). Qwen3-8B Re-DocRED cells use $n{=}949$.}
\label{tab:tracka-docred}
\centering\footnotesize
\begin{tabular}{llcccccc}
\toprule
Backbone & Config & F1 & Ign F1 & Cons & $T$/$V$ & ZT\% & F1 (Re-DocRED) \\
\midrule
GPT-5.2 & base & .2976 & .2886 & .0951 & 347/314 & 84.67 & .2855 \\
 & cot & .2653 & .2637 & .3716 & 549/345 & 78.46 & .2463 \\
 & cues-only & .3042 & .2945 & .5879 & 1075/443 & 62.63 & .2881 \\
 & sr & .2939 & .2815 & 1.0000 & 1300/0 & 49.40 & \textbf{.2960} \\
 & cot+sr & .2741 & .2704 & .9936 & 1086/7 & 62.32 & .2492 \\
 & Self-Refine & \textbf{.3289} & .3274 & .0728 & 302/280 & 85.27 & .2721 \\
 & Reflexion & .3270 & .3241 & .0898 & 256/233 & 86.87 & .2768 \\
 & CoVe & .2873 & .2820 & .0482 & 332/316 & 85.17 & .2562 \\
\midrule
Qwen3-8B & base 0-shot & .1612 & .1672 & .0194 & 772/757 & 67.03 & .1476 \\
 & cot+sr 0-shot & .1645 & .1698 & .9330 & 1223/82 & 65.03 & .1512 \\
 & base 5-shot & .1725 & .1767 & .0309 & 711/689 & 70.34 & .1684 \\
 & cot 5-shot & .1790 & .1855 & .0369 & 732/705 & 65.23 & .1618 \\
 & sr 5-shot & .1840 & .1897 & .9218 & 1240/97 & 63.53 & .1651 \\
 & cot+sr 5-shot & .1819 & .1864 & .9524 & 1071/51 & 60.52 & .1659 \\
\bottomrule
\end{tabular}
\end{table*}

\paragraph{Benchmarks and the paired protocol.}
\label{sec:setup:data}
\label{sec:setup:dual}
We evaluate on the full DocRED dev split (998 documents)~\cite{yao2019docred} and on DWIE~\cite{zaporojets2021dwie}, whose 200 dev+test documents are used only for final evaluation. Every prediction set on the DocRED documents is scored twice: against the original gold, and against Re-DocRED's revised annotations of the same 998 documents~\cite{tan2022revisiting}. Rescoring requires no new inference and leaves the gold label as the only variable. Therefore, any ranking change comes from the annotation. Re-DocRED also revises some entities; see Limitations. Track~B trains on 1{,}189 DocRED and 598 DWIE training documents (Appendix~\ref{app:training}).

\paragraph{Metrics.}
\label{sec:setup:eval}
We report micro-F1 and Ign F1, which drops gold triples that also appear in the training split; unlike the official Ign F1, which adjusts only precision, it can exceed F1. Unresolved entities, illegal labels, and self-loops count as false positives rather than being dropped. Cons is reported before post-processing, after which it is 1.0 by construction, together with the trigger counts $T$/$V$ and the zero-trigger rate (details in Appendix~\ref{app:eval}).

\paragraph{Systems.}
\label{sec:setup:systems}
Track~A runs GPT-5.2 through its API with provider-default decoding (the endpoint rejects user-set temperatures) and Qwen3-8B~\cite{yang2025qwen3} through vLLM~\cite{kwon2023efficient}. The Track~B students are built on Qwen3-8B and evaluated under transformers~\cite{wolf2020transformers}. We set $(\lambda_F, \lambda_C, \lambda_M) = (0.7, 0.2, 0.1)$ before training, carried over from preliminary experiments under an earlier evaluation protocol and not re-tuned: F1 carries most of the weight because it is the evaluation target, and the format term only rewards schema-valid labels (Appendix~\ref{app:grpo}). Our variants use five fixed teacher-trace demonstrations; the generic baselines (Self-Refine, Reflexion, CoVe) use the same demonstrations on DWIE and shell demonstrations built from gold on DocRED (Appendix~\ref{app:prompts}).

\paragraph{Statistics and reproducibility.}
\label{sec:setup:stats}
\label{sec:setup:repro}
All comparisons use document-level paired permutation tests (5{,}000 resamples). Test families were fixed before inspecting outcomes, and we report Bonferroni thresholds with the raw $p$-values (seven Track~A comparisons on DocRED, threshold 0.0071; twelve Track~B tests, 0.004; Appendix~\ref{app:stats}). Code, prompts, rule files, and run records are released (Appendix~\ref{app:repro}).

\section{Results}
\label{sec:results}

\subsection{DocRED: Same Predictions, Reversed Ranking}
\label{sec:results:docred}

\begin{figure}[t]
\centering
\includegraphics[width=\columnwidth]{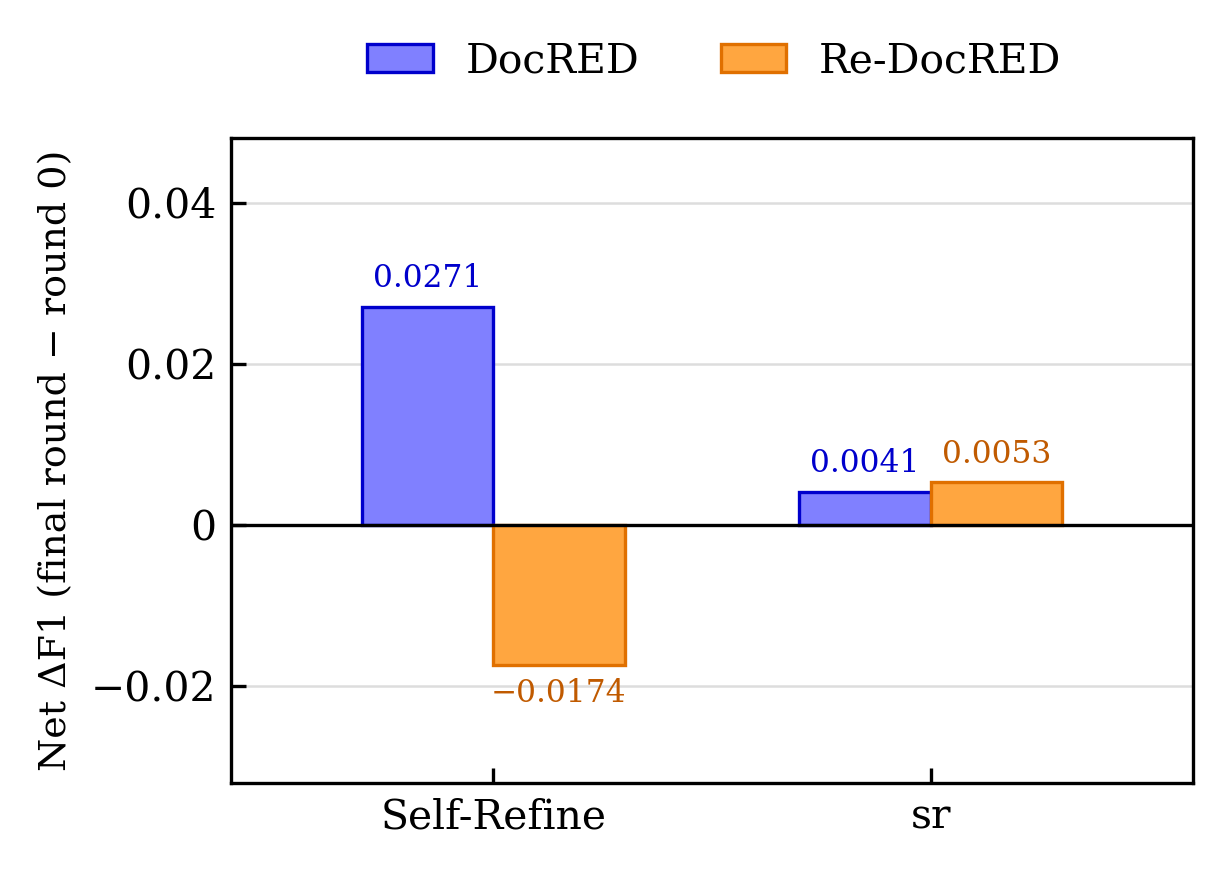}
\caption{Net F1 change of the reflection loop (final round minus round 0) on identical predictions scored under DocRED and Re-DocRED.}
\label{fig:loopflip}
\end{figure}

Under the original DocRED annotations, generic reflection leads (Table~\ref{tab:tracka-docred}): Self-Refine reaches 0.329 micro-F1 against 0.298 for \textsc{base}, and the constraint-feedback variant \textsc{sr} trails it by $0.035$ ($p{=}0.0002$, Bonferroni-passed; Appendix~\ref{app:stats}). Rescoring the \emph{identical} predictions against Re-DocRED reverses the ranking: \textsc{sr} moves to first, $0.019$ above the strongest baseline (Reflexion; $p{=}0.0002$), although its smaller leads over \textsc{base} and \textsc{cues-only} miss the Bonferroni threshold ($p{=}0.011$, $0.033$; Appendix~\ref{app:stats}). The loop view is starker (Figure~\ref{fig:loopflip}): the net effect of the Self-Refine loop changes sign ($+0.0271\rightarrow-0.0174$), while the constraint-feedback loop is stable ($+0.0041\rightarrow+0.0053$).

Decomposing each loop's net effect into edit components (Appendix~\ref{app:editdecomp}) locates the flip. Self-Refine deletes 6{,}942 predictions: 14\% are true relations under DocRED but 31\% under Re-DocRED, so its deletion component flips from $+0.009$ to $-0.051$, while its additions grow from $+0.005$ to $+0.028$. The largest term under DocRED is neither: 48\% of the $+0.027$ comes from fewer unresolvable mentions (1{,}629$\rightarrow$575), a formatting effect of the second pass rather than better extraction. Constraint loops are free of this confound (unresolved component at most $+0.0015$), and the gold alone decides what their edits are worth: the same 364 feedback-named additions of \textsc{sr} hit gold 13.7\% of the time under DocRED and 43.1\% under Re-DocRED.

\paragraph{Backbone.}
\label{sec:results:backbone}
On Qwen3-8B (Table~\ref{tab:tracka-docred}, lower block), the same pipeline raises Cons from 0.02 to 0.95, while F1 changes by at most 0.012, and in the 0-shot setting only 10.5\% of closure additions hit gold. Structural repair transfers across backbones, but closure is only as precise as the predicted triples it starts from.

\paragraph{Cues and enforcement.}
Ablating the two layers separately on DocRED (Appendix~\ref{app:ablation}) shows that they act independently: without inverse-relation cues the model builds far less structure ($T$ $1075\rightarrow477$) at flat F1, and disabling enforcement is near-F1-neutral.

\subsection{DWIE: Where Constraint Feedback Pays}
\label{sec:results:dwie}

\begin{table}[t]
\caption{Track A on DWIE (GPT-5.2, 200 dev+test documents; post-processing off unless marked). Post-processed Cons is 1.0 by construction and not reported.}
\label{tab:dwie}
\centering\small
\begin{tabular}{lcc}
\toprule
Config & F1 & Cons \\
\midrule
base & .6711 & .454 \\
base + post-processing & \textbf{.7109} & -- \\
cot & .6635 & .505 \\
sr & .6437 & 1.000 \\
cot+sr & .6944 & .993 \\
Self-Refine & .6466 & -- \\
Reflexion & .6482 & -- \\
CoVe & .5875 & -- \\
\bottomrule
\end{tabular}
\end{table}

\begin{figure*}[t]
\centering
\includegraphics[width=\textwidth]{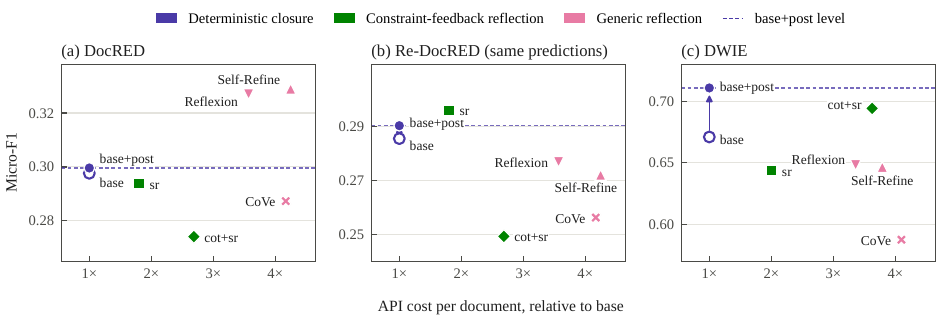}
\caption{Cost and micro-F1 of Track~A configurations under three golds (GPT-5.2). The $x$-axis is API cost per document relative to \textsc{base}, at identical token prices for all configurations; post-processing makes no API call, so \textsc{base} and \textsc{base}+post share a cost. Panels (a) and (b) score the same predictions against DocRED and Re-DocRED. Seed 20261012; $y$-axes differ across panels.}
\label{fig:cost}
\end{figure*}

On DWIE, whose gold is near-closed, the same mechanisms pay (Table~\ref{tab:dwie}). \textsc{cot+sr} beats all three generic baselines by $+0.046$/$+0.048$/$+0.107$ micro-F1 (all $p{=}0.0002$, Bonferroni-passed), about $18\times$ its three-seed standard deviation (0.0025). The constraint fields tax the first pass: \textsc{sr} opens 0.047 below \textsc{base} (0.624 vs.\ 0.671) and its loop recovers only part of the gap, so \textsc{sr} alone is indistinguishable from Reflexion and Self-Refine ($p{=}0.72$, $0.82$), whereas \textsc{cot+sr} opens near \textsc{base} (0.668) and gains $+0.026$ in the loop. The deterministic layer alone does better still: \textsc{base} with post-processing reaches 0.710 against 0.692 for \textsc{cot+sr} (three-seed means) at $3.6\times$ lower cost (\$2.16 vs.\ \$7.85 per run; Figure~\ref{fig:cost}); the gap holds in every seed but misses the Bonferroni threshold ($p{=}0.037$). Reflection without constraint feedback hurts here: Self-Refine's loop deletes 989 predictions, 48\% of them true, and loses 0.026 F1---the same deletions that DocRED's gaps reward.

\subsection{Across Golds: Closure Predicts Repair Precision}
\label{sec:results:synthesis}

\begin{table}[t]
\caption{Edit quality under three golds. The DocRED and Re-DocRED columns score identical predictions; DWIE is near-closed under its own rules. Row 1: hit rate of closure additions divided by the model's own precision (GPT-5.2, configurations without reflection); row 2: hit rate of the additions named by constraint feedback (\textsc{sr}); rows 3--5: Self-Refine's share of deletions that are gold, hit rate of its unprompted additions, and loop net effect.}
\label{tab:synthesis}
\centering\small
\resizebox{\columnwidth}{!}{%
\begin{tabular}{lccc}
\toprule
Measurement & DocRED & Re-DocRED & DWIE \\
\midrule
Closure additions: hit rate / P & 0.72 & 0.99 & 0.99 \\
Feedback-named additions: hit rate & .137 & .431 & .793 \\
Self-Refine deletions: gold share & 14.1\% & 31.2\% & 47.7\% \\
Self-Refine unprompted additions: hit rate & .180 & .388 & .524 \\
Self-Refine loop $\Delta$F1 & $+$.0271 & $-$.0174 & $-$.0257 \\
\bottomrule
\end{tabular}}
\end{table}

Table~\ref{tab:synthesis} lines up five measurements of edit quality. On identical predictions, the repaired gold credits additions more and generic deletions less, and DWIE, near-closed under its own rules, sides with Re-DocRED on every row. Deterministic closure adds facts that are as precise as the model's own predictions where the gold is closed (hit rate over model precision 0.99) but only about 70\% as precise under DocRED, and this discount is the same for both backbones (0.72 for GPT-5.2, 0.68 for Qwen3-8B) although their precision differs 1.6-fold: it is set by the gold, not by the model. Generic reflection mirrors this on the deletion side: the more complete the gold, the more of its deletions are true relations, and its loop gain survives only under DocRED.

\paragraph{Cost.} Figure~\ref{fig:cost} places the configurations on the cost--F1 plane. Generic reflection costs 3.4--4.2 times as much as \textsc{base} per document, and its F1 advantage exists only under DocRED: rescored against Re-DocRED, and on DWIE, it falls below even the single-pass \textsc{base} output. Deterministic closure adds to \textsc{base} at no API cost ($+0.005$ under Re-DocRED, $+0.040$ on DWIE). Where the gold is near-closed, the closed \textsc{base} output is therefore the baseline to beat before adding an LLM reflection loop.

\subsection{Training Time: Field Reversal and GRPO Gains}
\label{sec:results:trackb}

Figure~\ref{fig:trackb} summarizes Track~B; full results and further analysis are in Appendix~\ref{app:trackb}. Fewer verbalized fields train better students: \emph{gold-only} beats \emph{no-cons-field}, which beats \emph{full} traces, on both benchmarks (gold-only $-$ full: $+0.075$ on DocRED; $+0.183$ on DWIE, $+0.143$ after excluding the 12 documents the full student fails to parse; all $p{=}0.0002$).

GRPO on the gold-only student adds $+0.025$ on DocRED and $+0.019$ on DWIE (both $p{=}0.0002$), but the gains differ in kind. On DocRED, GRPO cuts predictions by 18\% and loses 0.018 recall; about 20\% of the predictions it prunes are true relations under Re-DocRED, and rescored against Re-DocRED the gain disappears ($-0.005$, $p{=}0.16$). On DWIE, predictions shrink by only 1.7\% and precision and recall rise together.

\begin{figure}[t]
\centering
\includegraphics[width=\columnwidth]{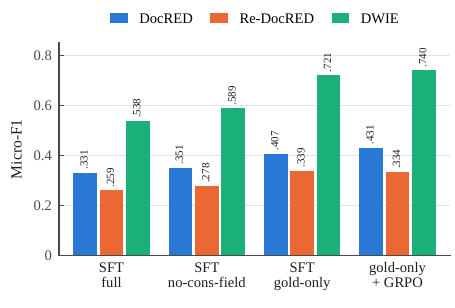}
\caption{Track~B students (Qwen3-8B): micro-F1 of the three SFT trace variants and of GRPO on the gold-only student (mean of three seeds); DocRED and Re-DocRED score the same predictions.  Cons, the full-trace GRPO runs, and seed spreads are in Table~\ref{tab:trackb}; DocRED values are transcribed from experiment logs (Limitations).}
\label{fig:trackb}
\end{figure}

\section{Discussion}
\label{sec:discussion}

\paragraph{A pre-deployment decision rule.}
The results reduce to a procedure that runs before any model call: audit the gold. If it is near-closed (DWIE-like), constraint feedback is worth its cost, and deterministic post-processing is the first component to deploy. If it is leaky (DocRED-like), apparent gains of generic reflection should be discounted: under repaired gold their deletion component flips sign, and for Self-Refine the largest component is a formatting effect. Constraint loops are roughly F1-neutral there but still buy internal coherence for downstream consumers. 

\paragraph{Evaluation practice.}
Comparing refinement methods on DocRED's original annotation alone is not a neutral default: the same predictions rank in the opposite order under repaired gold, and recent refinement frameworks~\cite{wang2026document,han2025cognitive,wang2025leveraging} report F1 without auditing the gold they are scored against. Two conventions would make such conclusions robust at near-zero cost: release per-document predictions so that they can be rescored under revised annotations, and report a closure audit alongside F1. The audit rules are best derived from the schema rather than mined from annotations, whose conventions a miner would read as rule confidence (Appendix~\ref{app:related}).

\section{Conclusion}
\label{sec:conclusion}

Consistency repair in LLM-based DocRE pays off where the gold is closed under the constraints it enforces, and $\Delta_{\mathrm{closure}}$, one deterministic audit of the gold, measures this before any model is run. One formulation, one enumerator, and one audit carried this conclusion across two benchmarks with different schemas, inference- and training-time tracks, and four mechanism families. In practice, where the audit shows the gold to be near-closed, the base output closed under the rules is the baseline to beat before adding an LLM reflection loop. Natural next steps are Re$^2$-DocRED~\cite{heng-etal-2026-re2} as a third annotation version, stronger RL variants such as DAPO~\cite{yu2026dapo}, and closure over entity-level properties such as types and aliases.

\section*{Limitations}

\begin{itemize}
\item \textbf{Recovered DocRED run records.} The DocRED rows of
Table~\ref{tab:trackb} were transcribed from experiment logs after a
hardware failure destroyed the run records; the DWIE-side records are
complete, and the affected cells are footnoted in the table. Separately,
49 Qwen3-8B documents were lost to a local unpacking failure; the
affected cells and paired analyses use $n{=}949$
(Table~\ref{tab:tracka-docred}).

\item \textbf{Shared SFT checkpoint.} The three GRPO seeds of each
variant share a single SFT run, so the reported seed variance covers the
RL stage only and does not include SFT-stage variance.

\item \textbf{Prompt sensitivity.} Each configuration uses one fixed
prompt phrasing; we did not measure sensitivity to rephrasing.

\item \textbf{Reward weights.} The GRPO reward weights were carried
over from preliminary experiments under an earlier evaluation protocol
and were not swept under the current one (Section~\ref{sec:setup:systems}).

\item \textbf{Demonstration asymmetry.} On DocRED the three
generic-reflection baselines use shell demonstrations while our variants
use teacher-trace demonstrations (Appendix~\ref{app:prompts}); the bias
works against our variants, so the baselines' advantage there is
conservative. On DWIE all systems use teacher-trace demonstrations, so
the baselines' disadvantage is not attributable to demonstration
quality.

\item \textbf{Shared DWIE training pool.} DWIE SFT and GRPO consume the
same 602 documents---a design concession---and 4 of the 602 were
excluded (3 overlength, 1 incomplete teacher output).

\item \textbf{Generation caps.} GRPO rollouts are capped at 1024 tokens;
8\% of SFT targets exceed this cap and about 5.9\% of rollouts fail to
parse. Evaluation decoding uses a 3072-token cap and is unaffected.

\item \textbf{Entity-resolution costs.} \texttt{index\_raw\_spans}
renders 20 surface forms ambiguous, costing 0.0009 F1; we accepted the
cost (Appendix~\ref{app:eval}). The hierarchical evidence bucket used in
the functional-constraint analysis inherits the gold type hierarchy's
incompleteness and overestimates clean violations; the bias direction is
conservative.

\item \textbf{Ablation resolution.} The per-family reward ablation (B8)
triggers composition-class rules only 61--72 times per 200 documents;
its null result reflects trigger volume, not seed count.

\item \textbf{Sampling implementations.} Track A runs Qwen3-8B under
vLLM 0.11.2 while Track B evaluates under transformers; sampling
behavior differs slightly between the two stacks.

\item \textbf{Cross-track comparability.} Track B trains on benchmark
data and Track A does not, so their comparison is not head-to-head; we
report it only as evidence that, when annotations are available,
training consistency into parameters is more effective than repairing at
inference time.

\item \textbf{Paired-protocol confound.} Re-DocRED revises entity
annotations as well as relations, so the dual-annotation comparison
includes an entity-side component. The direction of this component is
benign for our conclusions, but we report it rather than adjust for it.

\item \textbf{What F1 credits.} Part of every reflection loop's F1 gain
is a formatting component---unresolvable mentions dropping from 1{,}629
to 575 in the largest case (Self-Refine on DocRED)---rather than better
extraction. We report the full decomposition (Section~\ref{sec:results:docred},
Appendix~\ref{app:editdecomp}), but the headline F1 still credits this
component.

\item \textbf{Human evaluation.} We report no human evaluation. Cons is a deterministic rule-based measurement whose violations can be checked directly against the rule files, so human judgment is not load-bearing for any claim in this paper.

\end{itemize}

\bibliography{custom}

\appendix

\section{Complete Rule Tables and Grounding Evidence}
\label{app:rules}

\subsection{Provenance}
\label{app:provenance}
Table~\ref{tab:provenance} traces every rule set to its source. The two
benchmarks differ in where positive rules come from because their
schemas differ: Wikidata declarations are generic knowledge-base
constraints rather than annotation guidelines, so DocRED candidates are
screened against the gold annotations, whereas DWIE's positive rules are
the ones its authors specified alongside the annotation scheme.
Functional constraints are treated alike on both benchmarks: they follow
declared or written single-value semantics and are never screened by
satisfaction rates.

\begin{table*}[t]
\caption{Rule provenance. Positive rules come from external declarations; functional constraints follow declared or written single-value semantics on both benchmarks. All rule sets were fixed before any experiment reported here, and no rule was selected using model predictions or model performance: admission statistics are computed on gold annotations only.}
\label{tab:provenance}
\centering\small
\begin{tabular}{@{}llp{45mm}p{47mm}p{21mm}@{}}
\toprule
Benchmark & Family & Source and candidates & Admission & Final \\
\midrule
DocRED & positive & Wikidata declarations for all 96 relation types (snapshot 2026-08-26): 103 candidates (3 Sym, 20 Inv, 7 Comp, 73 Imp) & $\geq$20 triggers on dev and satisfaction $\geq$0.50 on both dev and train & 10 (2 Sym, 7 Inv, 1 Imp) \\
DocRED & functional & Wikidata single-value declarations: 18 & all kept, tiered by their wording: 3 hard (enforced), 15 soft (reported only) & 18 \\
DWIE & positive & the dataset authors' rules \citep[Appendix C]{zaporojets2021dwie}: 41 & drop the 16 rules over the 12 \texttt{-x} types, which leave the label space & 25 (2 Sym, 3 Inv, 13 Imp, 7 Comp) \\
DWIE & functional & ours (DWIE declares none): 23 candidate relation types & written at-most-one criterion, one decision and reason per relation (Table~\ref{tab:dwie-func}) & 10 \\
\bottomrule
\end{tabular}
\end{table*}

\subsection{DocRED: Derivation and Admission}
All 96 DocRED relation types were expanded into candidate rules from
Wikidata property declarations~\cite{vrandevcic2014wikidata} (snapshot
2026-08-26): property constraints (P2302), instance-of statements (P31),
inverse-property statements (P1696), and subproperty-of statements
(P1647). The candidate pool before screening: 3 Sym, 20 Inv, 7 Comp,
47 direct Imp plus 26 chained Imp, 3 hard and 15 soft single-value
declarations, and 6 antisymmetry declarations. Admission requires at
least 20 triggered instantiations on dev and a satisfaction rate of at
least 0.50 on \emph{both} dev and train\_annotated; the dual-split
requirement removes rules that fit dev alone. Ten positive rules pass
(2 Sym, 7 Inv, 1 Imp, 0 Comp). Table~\ref{tab:rules-docred} lists each
admitted rule with both satisfaction rates and its dev audit counts; the
rule file records each rule's source declaration and, for every
rejected candidate, the declaration it came from and the measured rates
that failed admission.

\begin{table*}[t]
\caption{Admitted DocRED rules with grounding evidence. $T$/$V$ and Cons are the 998-document dev audit; \textit{sat.} is the satisfaction rate on dev / train\_annotated. Per-rule source declarations are in \texttt{docred\_rules\_v1\_1.json}.}
\label{tab:rules-docred}
\centering\footnotesize
\setlength{\tabcolsep}{3pt}
\begin{tabular}{@{}p{7mm}p{68mm}p{9mm}p{9mm}p{13mm}p{31mm}@{}}
\toprule
ID & Rule (body $\Rightarrow$ head) & \multicolumn{1}{r}{$T$} & \multicolumn{1}{r}{$V$} & \multicolumn{1}{r}{Cons} & \multicolumn{1}{r}{sat. dev/train} \\
\midrule
D.1 & spouse & \multicolumn{1}{r}{82} & \multicolumn{1}{r}{6} & \multicolumn{1}{r}{0.9268} & \multicolumn{1}{r}{0.927/0.871} \\
D.2 & sibling & \multicolumn{1}{r}{134} & \multicolumn{1}{r}{16} & \multicolumn{1}{r}{0.8806} & \multicolumn{1}{r}{0.881/0.842} \\
D.3 & contains admin. territory $\Rightarrow$ located in admin. territory & \multicolumn{1}{r}{618} & \multicolumn{1}{r}{114} & \multicolumn{1}{r}{0.8155} & \multicolumn{1}{r}{0.816/0.803} \\
D.4 & characters $\Rightarrow$ present in work & \multicolumn{1}{r}{74} & \multicolumn{1}{r}{19} & \multicolumn{1}{r}{0.7432} & \multicolumn{1}{r}{0.743/0.528} \\
D.5 & follows $\Rightarrow$ followed by & \multicolumn{1}{r}{69} & \multicolumn{1}{r}{23} & \multicolumn{1}{r}{0.6667} & \multicolumn{1}{r}{0.667/0.793} \\
D.6 & followed by $\Rightarrow$ follows & \multicolumn{1}{r}{51} & \multicolumn{1}{r}{5} & \multicolumn{1}{r}{0.9020} & \multicolumn{1}{r}{0.902/0.776} \\
D.7 & father $\Rightarrow$ child & \multicolumn{1}{r}{59} & \multicolumn{1}{r}{8} & \multicolumn{1}{r}{0.8644} & \multicolumn{1}{r}{0.864/0.897} \\
D.10 & capital $\Rightarrow$ capital of & \multicolumn{1}{r}{27} & \multicolumn{1}{r}{8} & \multicolumn{1}{r}{0.7037} & \multicolumn{1}{r}{0.704/0.718} \\
D.11 & capital of $\Rightarrow$ capital & \multicolumn{1}{r}{21} & \multicolumn{1}{r}{2} & \multicolumn{1}{r}{0.9048} & \multicolumn{1}{r}{0.905/0.803} \\
D.12 & capital of $\Rightarrow$ located in the administrative territorial entity & \multicolumn{1}{r}{21} & \multicolumn{1}{r}{10} & \multicolumn{1}{r}{0.5238} & \multicolumn{1}{r}{0.524/0.539} \\
\midrule
\multicolumn{2}{@{}l}{\textbf{Total (10 positive rules)}} & \multicolumn{1}{r}{\textbf{1,156}} & \multicolumn{1}{r}{\textbf{211}} & \multicolumn{1}{r}{\textbf{0.8175}} & \\
\bottomrule
\end{tabular}
\end{table*}

\subsection{DWIE: Author Rules after \texttt{-x} Removal}
DWIE's own rule appendix~\cite{zaporojets2021dwie} contains 41 rules;
removing the 12 \texttt{-x} nominal-variant relation types leaves 25
positive rules (2 Sym, 3 Inv, 13 Imp, 7 Comp). DWIE's Appendix~C contains
no functional constraint; the ten we use are selected by the protocol in
Appendix~\ref{app:dwiefunc}.
The \texttt{-x} removal is justified by functionality: for the 8
base/\texttt{-x} pairs, multi-tail rates differ by two orders of
magnitude (e.g., \texttt{in0} 0.1\% vs.\ \texttt{in0-x} 19.2\%), fanned
out by rules C.15/C.16 through \texttt{gpe0}; deduplicating such pairs
is entity normalization, not logical inference. A side benefit is that
binary \texttt{gpe0} no longer appears in any retained rule. Two of the
25 rules carry type guards (C.29, C.35, both Imp); both have zero
implicit-side triggers (implicit entities have empty tag sets) and zero
violations, so they contribute nothing to $\Delta_{\mathrm{closure}}$.
Table~\ref{tab:rules-dwie} lists all 35 constraints with their audit
counts.

\subsection{DWIE Functional Constraints: Selection Protocol}
\label{app:dwiefunc}
DWIE declares no functional constraint, so the ten we use are ours. They
were fixed in four steps before any experiment reported here; the rule
file released with this paper records the criterion, each decision with
its reason, and the date.
\begin{enumerate}[leftmargin=*,nosep]
\item \emph{Candidates.} The 23 of DWIE's 53 relation types for which
a single-valued reading is plausible.
\item \emph{Semantic criterion.} A relation is functional if, under
DWIE's annotation semantics, the head entity determines the tail entity
within a document. We applied it relation by relation and recorded a
reason for every decision (Table~\ref{tab:dwie-func}); the three levels
of \texttt{gpe} and of \texttt{in} are treated alike.
\item \emph{Gold check.} Audited after the list was fixed, the ten
constraints hold for 4{,}780 of 4{,}784 anchors on the full 802-document
corpus (Cons 0.9992; Table~\ref{tab:rules-dwie}). Gold multi-tail counts
had been recorded for 16 candidates and were inspected alongside the
decisions, but they do not determine them: of the 12 candidates with at most one gold conflict, six
were admitted and six excluded (\texttt{directed\_by}, with none, is
excluded because co-directed works exist).
\item \emph{Admission.} All ten were kept and are enforced by
deduplication, untiered.
\end{enumerate}
Their direct effect on the reported results is negligible. No deletion
in any reflection loop was licensed by a functional conflict
(Appendix~\ref{app:editdecomp}), and over the eight post-processed
GPT-5.2 runs on DWIE, functional enforcement deleted two predictions in
total, at most one per run. We did not ablate them as prompt cues on
DWIE; on DocRED, removing the functional cues changes F1 by $-0.002$
(Appendix~\ref{app:ablation}). Applying DocRED's admission thresholds
(20 triggers, satisfaction 0.50) to DWIE's positive rules on the full
corpus would remove none on satisfaction (the lowest is 0.84, C.13) and
would drop only the four with fewer than 20 triggers (C.26, C.31, C.36,
C.38); none of them records a gold violation, so
$\Delta_{\mathrm{closure}}$ is unchanged.

\begin{table}[h!]
\caption{DWIE functional-constraint decisions under the written
at-most-one criterion (Appendix~\ref{app:dwiefunc}).}
\label{tab:dwie-func}
\centering\footnotesize
\begin{tabular}{@{}p{29mm}p{5mm}p{34mm}@{}}
\toprule
Relations & & Reason \\
\midrule
\texttt{gpe0}, \texttt{gpe1}, \texttt{gpe2} & in & a nominal variant denotes one GPE \\
\texttt{in0}, \texttt{in1}, \texttt{in2} & in & a place lies in one country, region, or city \\
\texttt{minister\_of} & in & a minister serves one government \\
\texttt{head\_of\_state}, \texttt{head\_of\_gov}, \texttt{mayor\_of} & in & one office at a time \\
\midrule
\texttt{citizen\_of} & out & dual citizenship \\
\texttt{based\_in0}, \texttt{based\_in1}, \texttt{based\_in2} & out & multinational organizations \\
\texttt{head\_of} & out & one person can head several bodies \\
\texttt{member\_of} & out & many-to-many by design \\
\texttt{spouse\_of} & out & successive marriages in one article \\
\texttt{directed\_by} & out & co-directed works \\
\texttt{agent\_of} & out & superclass of several offices \\
\texttt{event\_in0} & out & multi-country events \\
\texttt{agency\_of}, \texttt{institution\_of}, \texttt{ministry\_of} & out & borderline; excluded to keep the criterion conservative \\
\bottomrule
\end{tabular}
\end{table}

\begin{table*}[!tp]
\caption{Retained DWIE rules with audit counts on the full 802-document corpus. Positive rules come from the dataset authors' Appendix C; the functional constraints are ours (Appendix~\ref{app:dwiefunc}) and are listed separately. Two implications carry an entity-type guard, applied before grounding: C.29 requires \texttt{gpe0} (the unary entity-type tag, not the binary relation) on $Y$, and C.35 requires \texttt{sport\_player} on $X$; both guards are exercised in the $T$ column. Each functional denial asserts at most one tail per anchor. Generated from \texttt{dwie\_rules\_v1.1.json} and \texttt{audit\_dwie.json}.}
\label{tab:rules-dwie}
\centering\footnotesize
\setlength{\tabcolsep}{3pt}
\begin{tabular}{@{}p{7mm}p{85mm}p{10mm}p{10mm}p{13mm}p{13mm}@{}}
\toprule
ID & Rule & \multicolumn{1}{r}{$T$} & \multicolumn{1}{r}{$V$} & \multicolumn{1}{r}{Cons} & \multicolumn{1}{r}{class} \\
\midrule
\multicolumn{6}{@{}l}{\textit{Positive rules (DWIE Appendix C, after $-x$ removal)}}\\
C.1 & spouse\_of symmetric & \multicolumn{1}{r}{55} & \multicolumn{1}{r}{0} & \multicolumn{1}{r}{1.0000} & \multicolumn{1}{r}{sym} \\
C.2 & vs symmetric & \multicolumn{1}{r}{281} & \multicolumn{1}{r}{0} & \multicolumn{1}{r}{1.0000} & \multicolumn{1}{r}{sym} \\
C.4 & won\_vs $\Rightarrow$ vs (swapped) & \multicolumn{1}{r}{61} & \multicolumn{1}{r}{0} & \multicolumn{1}{r}{1.0000} & \multicolumn{1}{r}{inv} \\
C.5 & child\_of $\Rightarrow$ parent\_of & \multicolumn{1}{r}{22} & \multicolumn{1}{r}{0} & \multicolumn{1}{r}{1.0000} & \multicolumn{1}{r}{inv} \\
C.6 & parent\_of $\Rightarrow$ child\_of & \multicolumn{1}{r}{22} & \multicolumn{1}{r}{0} & \multicolumn{1}{r}{1.0000} & \multicolumn{1}{r}{inv} \\
C.3 & won\_vs $\Rightarrow$ vs & \multicolumn{1}{r}{61} & \multicolumn{1}{r}{0} & \multicolumn{1}{r}{1.0000} & \multicolumn{1}{r}{imp} \\
C.7 & ministry\_of $\Rightarrow$ agency\_of & \multicolumn{1}{r}{81} & \multicolumn{1}{r}{0} & \multicolumn{1}{r}{1.0000} & \multicolumn{1}{r}{imp} \\
C.12 & minister\_of $\Rightarrow$ agent\_of & \multicolumn{1}{r}{234} & \multicolumn{1}{r}{20} & \multicolumn{1}{r}{0.9145} & \multicolumn{1}{r}{imp} \\
C.13 & head\_of\_gov $\Rightarrow$ agent\_of & \multicolumn{1}{r}{272} & \multicolumn{1}{r}{44} & \multicolumn{1}{r}{0.8382} & \multicolumn{1}{r}{imp} \\
C.14 & head\_of\_state $\Rightarrow$ agent\_of & \multicolumn{1}{r}{379} & \multicolumn{1}{r}{19} & \multicolumn{1}{r}{0.9499} & \multicolumn{1}{r}{imp} \\
C.32 & head\_of $\Rightarrow$ member\_of & \multicolumn{1}{r}{563} & \multicolumn{1}{r}{0} & \multicolumn{1}{r}{1.0000} & \multicolumn{1}{r}{imp} \\
C.33 & coach\_of $\Rightarrow$ member\_of & \multicolumn{1}{r}{65} & \multicolumn{1}{r}{0} & \multicolumn{1}{r}{1.0000} & \multicolumn{1}{r}{imp} \\
C.34 & spokesperson\_of $\Rightarrow$ member\_of & \multicolumn{1}{r}{39} & \multicolumn{1}{r}{0} & \multicolumn{1}{r}{1.0000} & \multicolumn{1}{r}{imp} \\
C.36 & mayor\_of $\Rightarrow$ head\_of\_gov & \multicolumn{1}{r}{10} & \multicolumn{1}{r}{0} & \multicolumn{1}{r}{1.0000} & \multicolumn{1}{r}{imp} \\
C.37 & directed\_by $\Rightarrow$ created\_by & \multicolumn{1}{r}{39} & \multicolumn{1}{r}{0} & \multicolumn{1}{r}{1.0000} & \multicolumn{1}{r}{imp} \\
C.39 & institution\_of $\Rightarrow$ part\_of & \multicolumn{1}{r}{104} & \multicolumn{1}{r}{0} & \multicolumn{1}{r}{1.0000} & \multicolumn{1}{r}{imp} \\
C.29 & agency\_of + gpe0\textless{}Y\textgreater{} $\Rightarrow$ based\_in0 & \multicolumn{1}{r}{411} & \multicolumn{1}{r}{0} & \multicolumn{1}{r}{1.0000} & \multicolumn{1}{r}{imp} \\
C.35 & member\_of + sport\_player\textless{}X\textgreater{} $\Rightarrow$ player\_of & \multicolumn{1}{r}{401} & \multicolumn{1}{r}{0} & \multicolumn{1}{r}{1.0000} & \multicolumn{1}{r}{imp} \\
C.25 & in2 $\circ$ in0 $\Rightarrow$ in0 & \multicolumn{1}{r}{157} & \multicolumn{1}{r}{0} & \multicolumn{1}{r}{1.0000} & \multicolumn{1}{r}{comp} \\
C.26 & in1 $\circ$ in0 $\Rightarrow$ in0 & \multicolumn{1}{r}{13} & \multicolumn{1}{r}{0} & \multicolumn{1}{r}{1.0000} & \multicolumn{1}{r}{comp} \\
C.27 & based\_in2 $\circ$ in0 $\Rightarrow$ based\_in0 & \multicolumn{1}{r}{183} & \multicolumn{1}{r}{0} & \multicolumn{1}{r}{1.0000} & \multicolumn{1}{r}{comp} \\
C.28 & based\_in1 $\circ$ in0 $\Rightarrow$ based\_in0 & \multicolumn{1}{r}{21} & \multicolumn{1}{r}{0} & \multicolumn{1}{r}{1.0000} & \multicolumn{1}{r}{comp} \\
C.30 & event\_in2 $\circ$ in0 $\Rightarrow$ event\_in0 & \multicolumn{1}{r}{36} & \multicolumn{1}{r}{0} & \multicolumn{1}{r}{1.0000} & \multicolumn{1}{r}{comp} \\
C.31 & event\_in1 $\circ$ in0 $\Rightarrow$ event\_in0 & \multicolumn{1}{r}{1} & \multicolumn{1}{r}{0} & \multicolumn{1}{r}{1.0000} & \multicolumn{1}{r}{comp} \\
C.38 & character\_in + played\_by $\Rightarrow$ plays\_in & \multicolumn{1}{r}{4} & \multicolumn{1}{r}{0} & \multicolumn{1}{r}{1.0000} & \multicolumn{1}{r}{comp} \\
\midrule
\multicolumn{2}{@{}l}{\textbf{Total (25 positive)}} & \multicolumn{1}{r}{\textbf{3,515}} & \multicolumn{1}{r}{\textbf{83}} & \multicolumn{1}{r}{\textbf{0.9764}} & \\
\midrule
\multicolumn{6}{@{}l}{\textit{Functional constraints (ours; DWIE defines none).}}\\
F.1 & gpe0 at-most-one & \multicolumn{1}{r}{1,568} & \multicolumn{1}{r}{0} & \multicolumn{1}{r}{1.0000} & \multicolumn{1}{r}{func} \\
F.2 & gpe1 at-most-one & \multicolumn{1}{r}{35} & \multicolumn{1}{r}{0} & \multicolumn{1}{r}{1.0000} & \multicolumn{1}{r}{func} \\
F.3 & gpe2 at-most-one & \multicolumn{1}{r}{13} & \multicolumn{1}{r}{0} & \multicolumn{1}{r}{1.0000} & \multicolumn{1}{r}{func} \\
F.4 & in0 at-most-one & \multicolumn{1}{r}{2,105} & \multicolumn{1}{r}{2} & \multicolumn{1}{r}{0.9990} & \multicolumn{1}{r}{func} \\
F.5 & in1 at-most-one & \multicolumn{1}{r}{13} & \multicolumn{1}{r}{0} & \multicolumn{1}{r}{1.0000} & \multicolumn{1}{r}{func} \\
F.6 & in2 at-most-one & \multicolumn{1}{r}{157} & \multicolumn{1}{r}{0} & \multicolumn{1}{r}{1.0000} & \multicolumn{1}{r}{func} \\
F.7 & minister\_of at-most-one & \multicolumn{1}{r}{234} & \multicolumn{1}{r}{0} & \multicolumn{1}{r}{1.0000} & \multicolumn{1}{r}{func} \\
F.8 & head\_of\_state at-most-one & \multicolumn{1}{r}{378} & \multicolumn{1}{r}{1} & \multicolumn{1}{r}{0.9974} & \multicolumn{1}{r}{func} \\
F.9 & head\_of\_gov at-most-one & \multicolumn{1}{r}{271} & \multicolumn{1}{r}{1} & \multicolumn{1}{r}{0.9963} & \multicolumn{1}{r}{func} \\
F.10 & mayor\_of at-most-one & \multicolumn{1}{r}{10} & \multicolumn{1}{r}{0} & \multicolumn{1}{r}{1.0000} & \multicolumn{1}{r}{func} \\
\midrule
\multicolumn{2}{@{}l}{\textbf{Total (10 functional)}} & \multicolumn{1}{r}{\textbf{4,784}} & \multicolumn{1}{r}{\textbf{4}} & \multicolumn{1}{r}{\textbf{0.9992}} & \\
\bottomrule
\end{tabular}
\end{table*}

\subsection{Is Re-DocRED's Revision Rule-Driven?}
\label{app:a11}
If Re-DocRED had been completed by applying rules like ours, the paired
comparison would be circular. We therefore count, on the same 998
documents, how many of the triples Re-DocRED adds are heads of our ten
admitted rules whose bodies hold in the Re-DocRED gold. Additions are
matched by display name, because Re-DocRED merges coreference clusters and
entity indices do not carry across the two annotations. DocRED has 12{,}275
gold triples on these documents and Re-DocRED 34{,}631; Re-DocRED adds
22{,}754 (and removes about 400, so the net gain is 22{,}356). Of the
additions, only 1{,}547 (6.80\%) are derivable by the rules; 93.2\% are
unrelated to them. The derivable ones fall in nine relations
(Table~\ref{tab:a11}): six are fully derivable and all six are symmetric
or inverse pairs, where annotating both directions makes each direction
derivable from the other---the expected result of careful relabeling, not
evidence of rule-based completion. None comes from multi-hop reasoning, since
no composition rule is admitted on DocRED. The relation with the most
additions, \texttt{P131} (6{,}649, 29\% of all additions), is only 10.8\%
derivable.

\begin{table}[h]
\caption{Re-DocRED additions derivable by the ten admitted DocRED rules,
per relation (998 dev documents). All other relations contribute none.}
\label{tab:a11}
\centering\small
\resizebox{\columnwidth}{!}{%
\begin{tabular}{lrrrl}
\toprule
Relation & derivable & added & share & rule class \\
\midrule
P131 located in admin.\ entity & 717 & 6{,}649 & 10.8\% & Inv, Imp \\
P3373 sibling & 148 & 148 & 100\% & Sym \\
P156 followed by & 132 & 132 & 100\% & Inv \\
P1441 present in work & 127 & 265 & 47.9\% & Inv \\
P155 follows & 114 & 114 & 100\% & Inv \\
P26 spouse & 114 & 114 & 100\% & Sym \\
P40 child & 107 & 157 & 68.2\% & Inv \\
P1376 capital of & 47 & 47 & 100\% & Inv \\
P36 capital & 41 & 41 & 100\% & Inv \\
\midrule
Total & 1{,}547 & 22{,}754 & 6.80\% & \\
\bottomrule
\end{tabular}}
\end{table}

\subsection{Functional Tiers}
DocRED functional constraints are tiered by declaration semantics. Hard
single-value declarations cover only three properties (P194, P585,
P1198) and are enforced by keep-first deduplication. The soft tier (15
properties, including P569/P570/P19/P20/P22/P25) follows Wikidata's own
qualification that multiple values may coexist with different precisions
and references; soft violations are recorded but never removed---on
DocRED, time-precision variants account for 100\% of multi-tail cases
for the five temporal properties, so deduplicating them would merge
``1953'' with ``March 1953'', which is entity normalization. DWIE's ten
functional constraints are untiered and deduplicated, since the criterion
admits only relations read as at most one-valued (Appendix~\ref{app:dwiefunc}). Dev audit: hard $T{=}87$,
Cons$_{\mathrm{clean}}$ 0.9540; soft $T{=}1{,}304$,
Cons$_{\mathrm{clean}}$ 0.9870.

\subsection{Antisymmetry Declarations Not Incorporated}
Wikidata declares six antisymmetry constraints over DocRED relations
(P131, P361, P40, P22, P279, P25). We did not incorporate them into the
five-category system: antisymmetry is a negative pairwise constraint
whose violation pattern (both directions present) overlaps with
functional enforcement in the post-processor, and none of the six
survives the dual-split satisfaction screen as a positive implication in
either direction. They are recorded in the rule file for completeness.

\section{Prompts and Demonstrations}
\label{app:prompts}

\subsection{Track A Prompt Structure}
All Track A configurations share one template with three blocks: (i) the
task instruction and the relation inventory for the benchmark; (ii) the
document with its entity list, in which entities are presented under
their disambiguated names (e.g., \texttt{Greek\#1}, \texttt{Greek\#10})
with the instruction to use these exact names; (iii) the output schema.
Configurations differ only in which fields the schema requests:
\textsc{base} requests the triple list; \textsc{cot} adds a
\texttt{reasoning} field; \textsc{sr} adds a \texttt{consistency\_check}
field; \textsc{cot+sr} requests both. For constraint-aware variants, a
declarative cue block renders the admitted rules in natural language
between blocks (ii) and (iii). In the reflection round, the enumerator's
output is rendered as structured feedback naming which rule fails to
close and which fact is missing (positive rules) or which facts conflict
(functional rules), and the model is asked to revise its triple set
against exactly this feedback. The full prompt strings are released in
the reproduction package (Appendix~\ref{app:repro}).

\subsection{Teacher Prompt (Track B)}
The teacher sees the gold annotations: the prompt presents the document,
the entity list, and a fenced JSON block of gold triples, and asks the
teacher to (i) write the evidence-based reasoning for the annotated
relations and (ii) check the annotated set against the constraints, using
a fenced enumeration block produced by the rule engine (trigger counts
and the concrete missing/conflicting triples per rule). The teacher
echoes the gold \texttt{relation} field verbatim (99.9\% exact echo on
DocRED, 100\% on DWIE); when the enumeration block reports a gold
violation, the teacher records it in \texttt{consistency\_check} but does
not modify the gold triples. Both fenced blocks are stripped when
building training pairs, so the student never sees gold or engine output
at inference time.

\subsection{Demonstrations}
The five 5-shot demonstrations are fixed in \texttt{\_doc\_ids.json}
(\texttt{demo\_ids}) and shared by the GPT-5.2 and Qwen3-8B grids; they
are teacher-trace examples containing the \texttt{reasoning} and
\texttt{consistency\_check} fields. The three generic-reflection
baselines on DocRED (A2) instead use shell demonstrations constructed
from gold, which lack these two fields; we kept them because the
baselines are compared among themselves under the same condition, and
the bias works against our variants (a rerun would have cost \$95). The
detection diagnostic for shell demonstrations is the prompt token count:
a shell-demo \textsc{cot+sr} prompt has 4{,}219 tokens, versus
6{,}629--6{,}904 with real demonstrations. On DWIE, all systems use
teacher-trace demonstrations. The A10 5-shot cells were re-run with the
same five demonstrations after the shell-demo defect was found
(\texttt{make\_demo5.py}, \$0.05); the 0-shot cells use no
demonstrations and were unaffected.

\subsection{Output Schema and Parsing}
Outputs are JSON arrays of triples under benchmark-specific relation
labels. Unresolvable entity names, illegal relation labels, self-loops,
and duplicates are never dropped silently: they count as false positives
and are tallied separately (Appendix~\ref{app:eval}). Documents whose
output fails JSON parsing entirely are scored with an empty prediction
set.

\section{Evaluation Protocol Details}
\label{app:eval}

\subsection{Entity Resolution}
Entity strings are resolved against three alias layers: the canonical
\texttt{concepts.text} (DWIE) or cluster name (DocRED), all mention
surface forms, and the \texttt{link} title; matching is exact after
normalization. The link layer is the only source for DWIE's 1{,}119
implicit entities (no mention, no text); excluding it would cap recall
by protocol. Resolution \emph{fails} rather than guesses when one
normalized surface form points to multiple entities within a layer:
this rejects 0.478\% of DWIE resolutions (117/24{,}469 at the document
layer, 2/13{,}084 at the link layer) and 1.041\% of DocRED resolutions
(225/21{,}615). Picking a candidate by mention count or first occurrence
would shift F1 silently and unauditably.

\paragraph{Disambiguation suffixes (DocRED).}
413 of 19{,}528 clusters (2.1\%) have names that do not uniquely address
themselves---typically a short mention of one cluster equals another
cluster's full name. We append a stable index suffix to such cluster
names. Without suffixes, even perfect predictions cap at F1 0.9717 and
recall 0.9450 (692 ambiguity failures); with suffixes the ceiling is
1.0. The residual 5.5\% is an annotation artifact, not an extraction
signal: DocRED's task definition judges relations over given clusters,
and unaddressable names measure a protocol defect. The same mechanism
is nearly inert on DWIE.

\paragraph{Raw spans as aliases.}
DocRED tokenizes \texttt{sents} but not \texttt{vertexSet.name};
601/26{,}141 mentions (2.30\%) differ across the two (e.g.,
\texttt{February 14 , 2014}), and 581 of them no longer match after
normalization. We register the exact span via \texttt{pos} offsets (no
heuristic de-tokenization). A model copying names from the text then
faces a ceiling of 0.9701 instead of 0.9498; the entity-list path is
unaffected (1.0). The known cost of \texttt{index\_raw\_spans}: 20
surface forms (12 of them tokenized dates such as
\texttt{august 18 , 1945}) become ambiguous, raising the DocRED
\textsc{cot+sr} unresolved count from 524 to 547 and costing 0.0009
F1; we accepted this in exchange for a single unified scorer across
tracks.

\subsection{False-Positive Accounting}
Unresolved entities, illegal relation labels, self-loops, and duplicate
triples are counted as false positives, never dropped, and tallied
separately. Dropping them would let a model inflate precision by
emitting garbage: adding 20 junk triples to otherwise perfect
predictions drops precision from 1.0 to 0.4444 under our accounting.
Unreachable entities (participating in gold relations but unaddressable
by any alias) are zero on both benchmarks, so the protocol caps no
model's recall.

\subsection{Functional-Confusion Bucketing}
Apparent functional violations are separated into three evidence tiers,
reported side by side rather than merged into a single ``cleaned''
number: (i) \emph{deterministic}---the conflicting tails share a mention
string or differ only in time precision; (ii) \emph{hierarchy}---the
tails are related by the dataset's own type hierarchy, whose
completeness bounds this tier (it overestimates clean violations, a
conservative direction); (iii) \emph{morphological}---string-heuristic
variants. On the paired 998 documents, the hierarchy bucket grows from
2 (DocRED) to 52 (Re-DocRED): the revision introduces more
single-value conflicts, and the separation is more sensitive under the
revised gold, so the rise in net violations (17 $\rightarrow$ 61) is a
conservative estimate.

\subsection{Post-Processing Order and Timing}
\label{app:postproc}
After the \emph{final} reflection round only, the deterministic post-processor runs
first positive-rule closure to fixpoint, then Func exactly once on the
closed set. The order is forced by the logic, not tuned: positive Horn
rules contain no negation, so closure can never introduce a contradiction;
the only possible conflict is a closure-added triple colliding with a
functional constraint, and running closure first lets Func adjudicate such
collisions deterministically, with no oscillation (deduplicating first
would be pointless---closure would simply re-add the deleted facts).
Closure additions are appended \emph{after} the model's own triples, so the
keep-first tie-break of Func deduplication never evicts a model-predicted
triple. We run post-processing once rather than every round for two
reasons: the reflector must see its own structural errors, not an
already-repaired set; and the post-processing on/off comparison
(Section~\ref{sec:results:dwie})
only cleanly measures the deterministic layer if it is confined to the end.
Func follows the declaration semantics of each
schema~\cite{vrandevcic2014wikidata}: on DocRED~\cite{yao2019docred}, three
hard constraints are deduplicated while fifteen soft ones are report-only
(the schema explicitly allows multiple values under differing source or
precision); on DWIE, the ten functional constraints we define
(Appendix~\ref{app:dwiefunc}) are deduplicated, untiered. Evidence buckets for apparent functional conflicts are described in the
previous subsection.

\subsection{Repetition Penalty Scan}
The distilled student falls into degenerate repetition on roughly a
third of documents (the \texttt{reasoning} string never closes). We
selected the repetition penalty on 50 dev documents:

\begin{table}[h]
\caption{Repetition-penalty scan (Qwen3-8B SFT student, 50 dev
documents, greedy). Left of the optimum, parsing failures dominate;
right of it, legitimate repeated mentions are suppressed (gold averages
13.7 triples per document). Evaluation uses $rp{=}1.05$; GRPO rollouts
use $rp{=}1.1$ because sampling needs stronger repetition control than
greedy decoding.}
\label{tab:rpscan}
\centering\small
\resizebox{\columnwidth}{!}{%
\begin{tabular}{lccccc}
\toprule
$rp$ & parse rate & micro-F1 & P & R & preds/doc \\
\midrule
1.00 & 64\% & 0.2491 & 0.308 & 0.209 & 13.2 \\
1.02 & 86\% & 0.3178 & 0.346 & 0.294 & 12.6 \\
\textbf{1.05} & \textbf{98\%} & \textbf{0.3670} & \textbf{0.406} & \textbf{0.335} & \textbf{11.1} \\
1.08 & 100\% & 0.3076 & 0.356 & 0.270 & 9.5 \\
1.12 & 100\% & 0.2934 & 0.432 & 0.222 & 6.8 \\
\bottomrule
\end{tabular}}
\end{table}

\subsection{Metrics}
P/R/F1 are derived from tp/fp/fn counts only; derived quantities are
never accepted as input by the scorer, and records lacking Ign F1 are
refused at dump time. Ign F1 recomputes F1 after removing gold triples
that also appear in the training split. Macro-F1 is demoted to an
auxiliary column; all headline numbers are micro-F1 on the full split.
Cons is reported before post-processing because post-processing
additions become bodies of further rules: the trigger count itself drifts
between the off/on settings (up to +39.5\% observed), so a Cons value
detached from its $T$/$V$ is not interpretable.

\subsection{Zero-Trigger Handling}
In audits and evaluation, a document triggering no rule contributes an
\emph{undefined} Cons (never 1.0), and the zero-trigger rate is reported
with the same positive-only denominator as Cons. In the GRPO reward,
the Cons term is \emph{neutralized} on zero-trigger prompts (the term is
removed and the remaining weights renormalized). The alternative pays
for gaming: on a document with 10 gold triples, adding two spurious
triples that trigger and satisfy a symmetry rule moves F1 from 1.0 to
0.909 (reward contribution 0.636) but Cons from 0 to 1.0 (contribution
0.2), for a total of 0.800 $\rightarrow$ 0.936---the reward would
encourage prediction inflation, the exact opposite of the paper's
thesis. Scoring zero-trigger documents as 1.0 fails the other way: it
hands the policy a free 0.2 for predicting less. After neutralization
the same probe scores 1.000 vs.\ 0.936, so gaming is unprofitable.

\section{Complete Statistical Results}
\label{app:stats}

All tests are document-level paired permutations (5{,}000 resamples,
seed 20261012) with bootstrap CIs, from one shared implementation
(\texttt{a\_analyze} / \texttt{b4\_dwie\_tests}); every test list was
fixed before looking at the results.

\subsection{Track A on DWIE (200 documents)}
Seven planned comparisons, Bonferroni threshold $0.05/7 =
0.0071$:

\begin{table}[h]
\caption{Track A DWIE paired tests (main seed). $\star$: passes
Bonferroni.}
\label{tab:dwie-tests}
\centering\small
\resizebox{\columnwidth}{!}{%
\begin{tabular}{lccc}
\toprule
comparison & $\Delta$F1 & $p$ & \\
\midrule
cot+sr $-$ Reflexion & +0.0461 & 0.0002 & $\star$ \\
cot+sr $-$ Self-refine & +0.0478 & 0.0002 & $\star$ \\
cot+sr $-$ Cove & +0.1068 & 0.0002 & $\star$ \\
cot+sr $-$ base & +0.0233 & 0.0078 & \\
sr $-$ base & $-$0.0273 & 0.0220 & \\
cot $-$ base & $-$0.0075 & 0.3671 & \\
base+post $-$ cot+sr & +0.0166 & 0.0366 & (holds at 3 seeds, Table~\ref{tab:dwie-seeds}) \\
\bottomrule
\end{tabular}}
\end{table}

Per-configuration F1 with 95\% CI: base .6711 [.649, .693], cot .6635,
sr .6437, cot+sr .6944 [.672, .716], base+post .7109 [.690, .732],
Reflexion .6482, Self-refine .6466, Cove .5875 (full table emitted from
\texttt{tracka\_dwie/a12\_tests.json}).
\textsc{sr} alone is not distinguishable from the generic baselines:
Reflexion $-$ \textsc{sr} $+0.0045$ ($p{=}0.72$), Self-Refine $-$
\textsc{sr} $+0.0029$ ($p{=}0.82$).

\begin{table}[h]
\caption{Three-seed repeats on DWIE (seeds 20261012 / 7 / 2024). The
base+post $-$ cot+sr gap (+0.0175) is about three times either standard
deviation; the cot+sr advantage over the strongest baseline (+0.046) is
18$\times$ its seed noise (0.0025).}
\label{tab:dwie-seeds}
\centering\small
\resizebox{\columnwidth}{!}{%
\begin{tabular}{lccc}
\toprule
config & three F1 values & mean $\pm$ std & cost per run \\
\midrule
cot+sr (off) & .6944 / .6895 / .6929 & 0.6923 $\pm$ 0.0025 & \$7.85 \\
base (off) & .6711 / .6763 / .6706 & 0.6727 $\pm$ 0.0032 & \$2.16 \\
base+post (on) & .7109 / .7147 / .7037 & 0.7098 $\pm$ 0.0056 & (same run) \\
\bottomrule
\end{tabular}}
\end{table}

\subsection{Track A on DocRED (998 documents)}
Key rows from the 16-configuration pairwise family
(\texttt{tracka\_docred/analysis\_w4.json}; the full matrix and all
bootstrap CIs are emitted by the analysis script):

\begin{table}[h]
\caption{Track A DocRED paired tests (post-processing off).}
\label{tab:docred-tests}
\centering\small
\resizebox{\columnwidth}{!}{%
\begin{tabular}{lcc}
\toprule
comparison & $\Delta$F1 & $p$ \\
\midrule
Self-refine $-$ cot+sr & +0.0548 & 0.0002 \\
Reflexion $-$ cot+sr & +0.0529 & 0.0002 \\
base $-$ cot+sr & +0.0235 & 0.0004 \\
base $-$ sr & +0.0038 & 0.4673 \\
cot+sr $-$ cot & +0.0088 & 0.0708 \\
\midrule
\multicolumn{3}{l}{\textbf{Planned family of seven (ranking reversal)}} \\
sr $-$ Self-Refine & $-$0.0351 & 0.0002$^\star$ \\
sr $-$ Reflexion & $-$0.0331 & 0.0002$^\star$ \\
sr $-$ cues-only & $-$0.0104 & 0.0248 \\
sr $-$ base & $-$0.0038 & 0.4673 \\
sr $-$ CoVe & +0.0066 & 0.2084 \\
cues-only $-$ Self-Refine & $-$0.0247 & 0.0002$^\star$ \\
cues-only $-$ Reflexion & $-$0.0228 & 0.0002$^\star$ \\
\bottomrule
\end{tabular}}
\end{table}

Qwen-side paired analyses use $n{=}949$ (49 documents lost to a local
unpacking failure); values differ from the 998-document records by at
most 0.002. The DocRED-side family for the ranking reversal (including
sr $-$ Self-Refine) was computed by the same script on \texttt{dev.json},
and both check anchors reproduce (sr $-$ base $p{=}0.47$; cues-only $-$
Self-Refine $p{=}0.0002$). On the identical predictions scored against
Re-DocRED (Table~\ref{tab:redocred-tests}), \textsc{sr} leads all three
generic baselines under the Bonferroni threshold, while its leads over
\textsc{base} and \textsc{cues-only} do not pass it.

\begin{table}[h]
\caption{Track A paired tests on Re-DocRED (identical predictions, 998
documents; the planned family of seven, Bonferroni threshold 0.0071).
$\star$: passes Bonferroni.}
\label{tab:redocred-tests}
\centering\small
\begin{tabular}{lcc}
\toprule
comparison & $\Delta$F1 & $p$ \\
\midrule
sr $-$ Reflexion & +0.0192 & 0.0002$^\star$ \\
sr $-$ Self-Refine & +0.0239 & 0.0002$^\star$ \\
sr $-$ CoVe & +0.0398 & 0.0002$^\star$ \\
sr $-$ base & +0.0105 & 0.0108 \\
sr $-$ cues-only & +0.0081 & 0.0330 \\
cues-only $-$ Self-Refine & +0.0158 & 0.0004$^\star$ \\
cues-only $-$ Reflexion & +0.0110 & 0.0106 \\
\bottomrule
\end{tabular}
\end{table}

\subsection{Constraint Ablation}
\label{app:ablation}
Table~\ref{tab:ablation} ablates the cue layer and the enforcement layer
independently. Removing inverse-relation cues collapses the structure the
model builds ($T$ $1075 \rightarrow 477$; Cons $0.588 \rightarrow 0.442$)
while F1 stays flat; disabling enforcement is near-F1-neutral (Inv
additions $451 \rightarrow 10$; Func deletions $41 \rightarrow 0$), and the
two paths do not interfere. On DocRED only the Inv and Func families carry
information---Comp rules failed admission and Imp triggers are negligible
(Section~\ref{sec:formal:grounding}). The two layers are therefore separable, and each is
measured on its own terms.

\begin{table}[t]
\caption{Constraint ablation on DocRED (998 documents), one arm per row:
cue layer (top) and enforcement layer (bottom). DocRED carries only the
Inv and Func families---Comp rules failed admission and Imp triggers are
negligible (Section~\ref{sec:formal:grounding}). Cue-arm $\Delta$F1 is measured against Full
with post-processing off, enforcement-arm $\Delta$F1 against Full with
post-processing on.}
\label{tab:ablation}
\centering\small
\resizebox{\columnwidth}{!}{%
\begin{tabular}{llccc}
\toprule
Benchmark & Arm & $\Delta T$ & $\Delta$Cons & $\Delta$F1 \\
\midrule
DocRED & $-$cue (Inv) & $1075{\rightarrow}477$ & $.588{\rightarrow}.442$ & $+.0010$ \\
DocRED & $-$cue (Func) & -- & $.926{\rightarrow}.858$ (hard) & $-.0020$ \\
DocRED & $-$enforce (Inv) & \multicolumn{2}{c}{additions $451{\rightarrow}10$} & $-.0013$ \\
DocRED & $-$enforce (Func) & \multicolumn{2}{c}{deletions $41{\rightarrow}0$} & $-.0005$ \\
\bottomrule
\end{tabular}}
\end{table}

\subsection{Track B Results}
\label{app:trackb}
Table~\ref{tab:trackb} gives the full Track~B results summarized in
Figure~\ref{fig:trackb}.

\begin{table}[h]
\caption{Track B (Qwen3-8B students). GRPO rows: three seeds sharing one SFT checkpoint, mean $\pm$ std, Cons as the seed range; the $1{\times}10^{-5}$ row is a single seed. DocRED values are transcribed from experiment logs after a hardware failure (Limitations). Re-DocRED applies to DocRED only.}
\label{tab:trackb}
\centering\small
\resizebox{\columnwidth}{!}{%
\begin{tabular}{llccc}
\toprule
Benchmark & Config & F1 & Cons & F1 (Re-DocRED) \\
\midrule
DocRED & baseline & .0136 & -- & .0094 \\
DocRED & SFT full & .3313 & .421 & .2594 \\
DocRED & SFT no-cons-field & .3514 & .397 & .2783 \\
DocRED & SFT gold-only & .4066 & .567 & .3392 \\
DocRED & SFT full + GRPO (lr $1{\times}10^{-5}$) & .3300 & -- & .2616 \\
DocRED & SFT full + GRPO (lr $5{\times}10^{-5}$) & $.3417 \pm .0036$ & .43--.45 & $.2730 \pm .0056$ \\
DocRED & SFT gold-only + GRPO (lr $5{\times}10^{-5}$) & $.4313 \pm .0053$ & .607--.667 & .3339 \\
\midrule
DWIE & baseline & .0829 & .182 & -- \\
DWIE & SFT full & .5384 & .690 & -- \\
DWIE & SFT no-cons-field & .5886 & .749 & -- \\
DWIE & SFT gold-only & .7214 & .909 & -- \\
DWIE & SFT gold-only + GRPO & $.7404 \pm .0041$ & .900 & -- \\
\bottomrule
\end{tabular}}
\end{table}

\subsection{Track B Paired Tests}
\begin{table}[h]
\caption{Track B paired tests. DocRED block: five planned
comparisons on 998 documents under both gold versions (12 tests were
run in total; the DocRED-side $p{=}0.024$ row does not pass the
Bonferroni threshold 0.004 and both $p$ values are reported). DWIE
block: four comparisons on 200 documents, all passing Bonferroni
($0.05/4 = 0.0125$). GRPO rows are seed-averaged tests fixed before
inspection; they do not cover SFT-stage variance (three seeds share one
SFT checkpoint).}
\label{tab:trackb-tests}
\centering\small
\resizebox{\columnwidth}{!}{%
\begin{tabular}{lcccc}
\toprule
 & \multicolumn{2}{c}{DocRED} & \multicolumn{2}{c}{Re-DocRED} \\
comparison & $\Delta$F1 & $p$ & $\Delta$F1 & $p$ \\
\midrule
gold-only $-$ full SFT & +0.0753 & 0.0002 & +0.0798 & 0.0002 \\
no-cons-field $-$ full SFT & +0.0202 & 0.0010 & +0.0190 & 0.0002 \\
GRPO $-$ SFT (full, 3-seed avg) & +0.0104 & 0.0236 & +0.0136 & 0.0008 \\
GRPO $-$ SFT (gold-only, 3-seed avg) & +0.0246 & 0.0002 & $-$0.0053 & 0.164 \\
GRPO v6 (lr $1{\times}10^{-5}$) $-$ SFT & $-$0.0013 & 0.81 & +0.0022 & 0.60 \\
\midrule
\multicolumn{5}{l}{\textbf{DWIE} (single gold)} \\
gold-only $-$ full SFT & +0.1831 & 0.0002 & \multicolumn{2}{c}{--} \\
no-cons-field $-$ full SFT & +0.0503 & 0.0002 & \multicolumn{2}{c}{--} \\
gold-only $-$ no-cons-field & +0.1328 & 0.0002 & \multicolumn{2}{c}{--} \\
GRPO $-$ gold-only SFT (3-seed avg) & +0.0190 & 0.0002 & \multicolumn{2}{c}{--} \\
\bottomrule
\end{tabular}}
\end{table}

Per-seed F1 (DocRED): full + GRPO 0.3449 / 0.3423 / 0.3378; gold-only +
GRPO 0.4319 / 0.4257 / 0.4363. Per-seed F1 (DWIE gold-only + GRPO):
0.7408 / 0.7361 / 0.7444, with per-seed tests s42 +0.0194 ($p{=}0.0008$),
s7 +0.0147 ($p{=}0.0098$), s2024 +0.0229 ($p{=}0.0002$).
Parse failures explain about 22\% of the DWIE field-reversal gap: on the
188 documents where both the full and gold-only students parse, gold-only
$-$ full is $+0.1430$, no-cons-field $-$ full $+0.0399$, and gold-only $-$
no-cons-field $+0.1031$ (all $p{=}0.0002$). Students trained on relations alone also reach the highest Cons on both benchmarks (DocRED 0.567 vs.\ 0.421 for full traces and 0.397 without the check field; DWIE 0.909 vs.\ 0.690 and 0.749; Table~\ref{tab:trackb}). The learning rate $1{\times}10^{-5}$ has no effect ($\Delta$F1 $-0.0013$, $p{=}0.81$; Table~\ref{tab:trackb-tests}), the Cons reward does not raise Cons on DWIE ($0.909\rightarrow0.900$), and a per-family reward ablation is too sparse to resolve anything (Table~\ref{tab:b8}). Because Track~B trains on the benchmark, the cross-track comparison says only that, when annotations are available, training consistency into parameters beats repairing outputs at inference time.

\subsection{Per-Class Trigger--Violation Report and the B8 Reward Ablation}
The full per-class ($T$/$V$/Cons/zero-trigger) report for every
configuration is emitted from the \texttt{by\_class} fields of the run
records (B5); Table~\ref{tab:b8} shows the constraint-type reward
ablation as one instance.

\begin{table}[h]
\caption{B8: removing one rule family from the GRPO reward's Cons term
(DWIE, seed 42, shared gold-only SFT). Neither ablated cell leaves the
full-reward three-seed Cons band (Comp [.970, 1.000]; Imp [.885,
.915]): Comp triggers only 61--72 times per 200 documents with 0--2
violations, so one violation moves Cons by 0.014---the ablation has no
resolution at this trigger volume. F1 effects are likewise null
($-$Comp $\Delta$F1 $-$0.0117, $p{=}0.120$; $-$Imp $+$0.0017,
$p{=}0.701$; seed std 0.0041).}
\label{tab:b8}
\centering\small
\resizebox{\columnwidth}{!}{%
\begin{tabular}{lccccc}
\toprule
config & micro-F1 & Comp $T/V{\rightarrow}$Cons & Imp $T/V{\rightarrow}$Cons & Inv & Sym \\
\midrule
gold-only SFT & 0.7214 & 70/2 $\rightarrow$ .971 & 633/53 $\rightarrow$ .916 & 29/6 $\rightarrow$ .793 & 44/10 $\rightarrow$ .773 \\
full reward s42 & 0.7408 & 67/2 $\rightarrow$ .970 & 616/71 $\rightarrow$ .885 & 26/5 $\rightarrow$ .808 & 34/6 $\rightarrow$ .824 \\
full reward s7 & 0.7361 & 69/2 $\rightarrow$ .971 & 611/53 $\rightarrow$ .913 & 25/5 $\rightarrow$ .800 & 27/7 $\rightarrow$ .741 \\
full reward s2024 & 0.7444 & 61/0 $\rightarrow$ 1.000 & 626/53 $\rightarrow$ .915 & 12/5 $\rightarrow$ .583 & 36/12 $\rightarrow$ .667 \\
$-$Comp & 0.7291 & 70/1 $\rightarrow$ \textbf{.986} & 591/59 $\rightarrow$ .900 & 22/8 $\rightarrow$ .636 & 39/17 $\rightarrow$ .564 \\
$-$Imp & 0.7425 & 72/2 $\rightarrow$ .972 & 588/69 $\rightarrow$ \textbf{.883} & 14/2 $\rightarrow$ .857 & 35/15 $\rightarrow$ .571 \\
\bottomrule
\end{tabular}}
\end{table}

\subsection{Closure-Addition Precision}
\label{app:closureprec}
Row~1 of Table~\ref{tab:synthesis} compares the hit rate of the triples
that deterministic closure adds (hits / (hits + misses); deletions are
excluded) with the precision of the model's own predictions. Only
configurations without reflection are used: after a reflection loop, few
violations remain and the residual additions are too few to score.
Where the gold is near-closed, closure additions are as precise as the
predictions they start from; under DocRED they lose about 30\% of that
precision, for both backbones, although the two backbones' precision
differs 1.6-fold. The lowest cell, Qwen3-8B 0-shot on DocRED, hits 80
of 765 additions (10.5\%).

\begin{table}[h]
\caption{Closure-addition hit rate vs.\ model precision (configurations
without reflection). Ranges span the configurations in each row.}
\label{tab:closureprec}
\centering\small
\resizebox{\columnwidth}{!}{%
\begin{tabular}{lcccc}
\toprule
Gold / backbone & cells & model P & hit rate & ratio \\
\midrule
DWIE / GPT-5.2 & 4 & .725--.807 & .677--.812 & 0.991 $\pm$ 0.038 \\
Re-DocRED / GPT-5.2 & 2 & .443--.462 & .432--.462 & 0.987 \\
DocRED / GPT-5.2 & 3 & .255--.281 & .186--.195 & 0.716 \\
DocRED / Qwen3-8B & 3 & .154--.175 & .105--.121 & 0.677 \\
\bottomrule
\end{tabular}}
\end{table}

\subsection{Edit Decomposition of Loop Net Effects}
\label{app:editdecomp}

Every reflection loop's net effect $\mathrm{F1}_f - \mathrm{F1}_0$
(round 0 $\rightarrow$ final) decomposes into five edit components:
licensed additions (feedback-named), unlicensed additions, unlicensed
deletions, licensed deletions (functional conflict groups), and the
unresolved component. The decomposition closes exactly: the five
components sum to the measured $\Delta$F1 with a maximum error of
$8.33{\times}10^{-17}$ across all 15 settings, and licensed deletions
are zero everywhere. All values are recomputed from per-document caches
(998 documents; 949 for Qwen), and the recomputed final F1 matches the
run records digit for digit. Loop net effects are reported on the
998-document subset throughout (the 1{,}000-document values for the
three generic baselines differ by at most 0.0002); only the 95\% CIs of
those baselines remain on the 1{,}000-document protocol.

\begin{table*}[t]
\caption{Round-0 vs.\ final accounting per setting. ``trig.'' counts
documents whose loop actually ran a second round (constraint loops
trigger only on violation, so their $\Delta$F1 is diluted over all
documents). $|P_0|/|P_f|$ for the Re-DocRED block are identical to the
DocRED block (same predictions) and repeated for readability.}
\label{tab:a15-accounting}
\centering\small
\resizebox{\textwidth}{!}{%
\begin{tabular}{lrrrrrrrccc}
\toprule
setting & $n$ & $|G|$ & $|P_0|$ & $|P_f|$ & TP$_0$ & TP$_f$ & F1$_0$ & F1$_f$ & $\Delta$F1 & trig. \\
\midrule
\multicolumn{11}{l}{\textbf{DocRED gold} (Qwen rows: $n{=}949$)} \\
Self-Refine & 998 & 12{,}275 & 15{,}825 & 10{,}447 & 4{,}241 & 3{,}737 & .3019 & .3289 & \textbf{+0.0271} & 998 \\
Reflexion & 998 & 12{,}275 & 15{,}619 & 10{,}877 & 4{,}193 & 3{,}785 & .3006 & .3270 & \textbf{+0.0263} & 998 \\
CoVe & 998 & 12{,}275 & 15{,}515 & 12{,}933 & 4{,}136 & 3{,}621 & .2977 & .2873 & $-$0.0104 & 998 \\
sr & 998 & 12{,}275 & 15{,}496 & 15{,}290 & 4{,}023 & 4{,}050 & .2897 & .2939 & +0.0041 & 295 \\
cot+sr & 998 & 12{,}275 & 10{,}723 & 10{,}643 & 3{,}127 & 3{,}141 & .2719 & .2741 & +0.0022 & 119 \\
Qwen sr 5-shot & 949 & 11{,}680 & 10{,}951 & 11{,}047 & 2{,}034 & 2{,}068 & .1798 & .1820 & +0.0022 & 307 \\
Qwen cot+sr 5-shot & 949 & 11{,}680 & 11{,}141 & 11{,}046 & 2{,}035 & 2{,}061 & .1783 & .1814 & +0.0030 & 309 \\
Qwen cot+sr 0-shot & 949 & 11{,}680 & 10{,}812 & 10{,}886 & 1{,}836 & 1{,}864 & .1633 & .1652 & +0.0019 & 302 \\
\midrule
\multicolumn{11}{l}{\textbf{Re-DocRED gold} (same predictions; Qwen $|G|{=}33{,}108$)} \\
Self-Refine & 998 & 34{,}631 & 15{,}825 & 10{,}447 & 7{,}302 & 6{,}132 & .2894 & .2721 & \textbf{$-$0.0174} & 998 \\
Reflexion & 998 & 34{,}631 & 15{,}619 & 10{,}877 & 7{,}281 & 6{,}299 & .2898 & .2768 & \textbf{$-$0.0130} & 998 \\
CoVe & 998 & 34{,}631 & 15{,}515 & 12{,}933 & 7{,}257 & 6{,}093 & .2894 & .2562 & $-$0.0332 & 998 \\
sr & 998 & 34{,}631 & 15{,}496 & 15{,}290 & 7{,}286 & 7{,}388 & .2907 & .2960 & +0.0053 & 295 \\
cot+sr & 998 & 34{,}631 & 10{,}723 & 10{,}643 & 5{,}589 & 5{,}642 & .2465 & .2492 & +0.0028 & 119 \\
Qwen sr 5-shot & 949 & 33{,}108 & 10{,}951 & 11{,}047 & 3{,}584 & 3{,}645 & .1627 & .1651 & +0.0024 & 307 \\
Qwen cot+sr 5-shot & 949 & 33{,}108 & 11{,}141 & 11{,}046 & 3{,}587 & 3{,}663 & .1621 & .1659 & +0.0038 & 309 \\
Qwen cot+sr 0-shot & 949 & 33{,}108 & 10{,}812 & 10{,}886 & 3{,}234 & 3{,}325 & .1473 & .1512 & +0.0039 & 302 \\
\midrule
\multicolumn{11}{l}{\textbf{DWIE gold} (200 documents)} \\
cot+sr & 200 & 3{,}624 & 2{,}492 & 2{,}586 & 2{,}044 & 2{,}156 & .6684 & .6944 & \textbf{+0.0260} & 99 \\
sr & 200 & 3{,}624 & 2{,}325 & 2{,}397 & 1{,}857 & 1{,}938 & .6243 & .6437 & \textbf{+0.0194} & 76 \\
Reflexion & 200 & 3{,}624 & 3{,}053 & 2{,}309 & 2{,}232 & 1{,}923 & .6686 & .6482 & $-$0.0203 & 200 \\
Self-Refine & 200 & 3{,}624 & 3{,}099 & 2{,}324 & 2{,}260 & 1{,}923 & .6723 & .6466 & $-$0.0257 & 200 \\
CoVe & 200 & 3{,}624 & 3{,}008 & 2{,}112 & 2{,}197 & 1{,}685 & .6625 & .5875 & $-$0.0750 & 200 \\
\bottomrule
\end{tabular}}
\end{table*}

\begin{table*}[t]
\caption{Five-component decomposition of $\Delta$F1. Each component
cell: edited triples / of which gold-true, contribution to $\Delta$F1.
Licensed deletions are zero in all 15 settings and omitted. The
unresolved component is the change in items counted in $|P|$ but not
resolvable to deduplicated triples (category (a), including (c)
duplicates, illegal labels, and self-loops; category (b) documents
whose parse status flips between rounds---CoVe 1, Reflexion 3,
Self-Refine 1, Qwen cot+sr 0-shot 18, Qwen sr 5-shot 18, Qwen cot+sr
5-shot 7---are counted normally inside the add/delete components).}
\label{tab:a15-decomp}
\centering\small
\resizebox{\textwidth}{!}{%
\begin{tabular}{lccccc}
\toprule
setting & $\Delta$F1 & licensed add & unlicensed add & unlicensed delete & unresolved \\
\midrule
\multicolumn{6}{l}{\textbf{DocRED gold}} \\
Self-Refine & +0.0271 & 5/3, +.0002 & 2{,}613/470, +.0046 & 6{,}942/977, +.0093 & \textbf{+.0131} \\
Reflexion & +0.0263 & 17/6, +.0003 & 2{,}210/389, +.0033 & 6{,}168/803, +.0129 & \textbf{+.0098} \\
CoVe & $-$0.0104 & 0/0, 0 & 40/8, +.0002 & 2{,}270/523, $-$.0144 & +.0039 \\
sr & +0.0041 & 359/50, $-$.0002 & 34/4, $-$.0001 & 453/27, +.0028 & +.0015 \\
cot+sr & +0.0022 & 132/37, +.0017 & 8/1, $-$.0000 & 209/24, +.0004 & +.0001 \\
Qwen sr 5-shot & +0.0022 & 436/42, +.0002 & 205/15, $-$.0003 & 503/23, +.0020 & +.0003 \\
Qwen cot+sr 5-shot & +0.0030 & 396/43, +.0006 & 254/28, +.0005 & 708/45, +.0016 & +.0003 \\
Qwen cot+sr 0-shot & +0.0019 & 422/34, $-$.0001 & 373/24, $-$.0006 & 629/30, +.0019 & +.0007 \\
\midrule
\multicolumn{6}{l}{\textbf{Re-DocRED gold (same predictions)}} \\
Self-Refine & $-$0.0174 & 6/6, +.0002 & 2{,}704/1{,}049, +.0280 & 7{,}121/2{,}225, $-$.0513 & +.0057 \\
Reflexion & $-$0.0130 & 17/14, +.0005 & 2{,}265/847, +.0220 & 6{,}329/1{,}843, $-$.0395 & +.0041 \\
CoVe & $-$0.0332 & 0/0, 0 & 40/14, +.0003 & 2{,}320/1{,}178, $-$.0353 & +.0017 \\
sr & +0.0053 & 359/155, +.0041 & 34/8, +.0001 & 461/61, +.0003 & +.0008 \\
cot+sr & +0.0028 & 132/87, +.0031 & 9/4, +.0001 & 209/38, $-$.0005 & +.0001 \\
Qwen sr 5-shot & +0.0024 & 432/92, +.0026 & 201/34, +.0008 & 487/65, $-$.0011 & +.0002 \\
Qwen cot+sr 5-shot & +0.0038 & 388/109, +.0035 & 254/75, +.0025 & 708/108, $-$.0023 & +.0001 \\
Qwen cot+sr 0-shot & +0.0039 & 411/103, +.0033 & 370/51, +.0011 & 627/63, $-$.0007 & +.0003 \\
\midrule
\multicolumn{6}{l}{\textbf{DWIE gold}} \\
cot+sr & +0.0260 & 167/143, +.0279 & 8/6, +.0011 & 78/37, $-$.0034 & +.0003 \\
sr & +0.0194 & 121/96, +.0193 & 1/0, $-$.0001 & 44/15, $-$.0004 & +.0006 \\
Reflexion & $-$0.0203 & 15/13, +.0026 & 185/111, +.0159 & 908/433, $-$.0425 & +.0038 \\
Self-Refine & $-$0.0257 & 14/14, +.0030 & 231/121, +.0142 & 989/472, $-$.0461 & +.0032 \\
CoVe & $-$0.0750 & 0/0, 0 & 6/5, +.0010 & 881/517, $-$.0782 & +.0021 \\
\bottomrule
\end{tabular}}
\end{table*}

Three prediction families fixed in advance all hold: 18/18 component
intervals contain the measured values; the DocRED$\rightarrow$Re-DocRED
swing is predicted to the third decimal for all three generic loops
(e.g., Self-Refine deletion swing predicted $-$0.059, measured
$-$0.0606; addition swing, including the unresolved component, +0.015 vs.\ +0.0160); addition magnitudes are
right on DWIE and underestimated by 35--40\% on DocRED (the estimate
used base's $|P_0|$ as a proxy). The same unlicensed additions are
recognized far more often under the repaired gold (Self-Refine 470
$\rightarrow$ 1{,}049 of 2{,}613/2{,}704; Reflexion 389 $\rightarrow$
847 of 2{,}210/2{,}265): unclosed gold depresses the measured hit rate
of additions as well as the measured quality of deletions. (The
Re-DocRED-side counts differ slightly because cluster merging changes
the index space; denominators are reported per row.)

\begin{table}[t]
\caption{Unlicensed additions split into rewrites (share an entity pair
with a deleted triple; label or direction changed) vs.\ new entity
pairs. Cells: triples / gold-true. On DocRED about one third of generic
loops' additions are rewrites; on DWIE, 9--18\%. Qwen3-8B rewrites over
half of its unlicensed additions (51--59\%), versus 15--25\% for
GPT-5.2---smaller models prefer relabeling to extracting anew.}
\label{tab:a15-rewrite}
\centering\small
\resizebox{\columnwidth}{!}{%
\begin{tabular}{lccc}
\toprule
setting & rewrite / hit & new pair / hit & rewrite share \\
\midrule
DocRED $\cdot$ Self-Refine & 959 / 170 & 1{,}654 / 300 & 36.7\% \\
DocRED $\cdot$ Reflexion & 753 / 129 & 1{,}457 / 260 & 34.1\% \\
DocRED $\cdot$ CoVe & 0 / 0 & 40 / 8 & 0.0\% \\
DocRED $\cdot$ Qwen cot+sr 5-shot & 151 / 18 & 103 / 10 & 59.4\% \\
DocRED $\cdot$ Qwen sr 5-shot & 108 / 5 & 97 / 10 & 52.7\% \\
DocRED $\cdot$ Qwen cot+sr 0-shot & 190 / 9 & 183 / 15 & 50.9\% \\
DWIE $\cdot$ Reflexion & 33 / 22 & 152 / 89 & 17.8\% \\
DWIE $\cdot$ Self-Refine & 21 / 14 & 210 / 107 & 9.1\% \\
DWIE $\cdot$ CoVe & 3 / 3 & 3 / 2 & 50.0\% \\
\bottomrule
\end{tabular}}
\end{table}

\begin{table}[t]
\caption{Two loop families, two benefit structures: component shares of
$\Delta$F1. Constraint loops' gains come almost entirely from the
feedback-named additions (over 100\% on DWIE, where their deletion
component is negative); the generic loop's DocRED gain is 0.7\%
feedback-named. These are not one mechanism at two strengths but two
mechanisms.}
\label{tab:a15-mech}
\centering\small
\resizebox{\columnwidth}{!}{%
\begin{tabular}{lccc}
\toprule
 & DWIE $\cdot$ cot+sr & DWIE $\cdot$ sr & DocRED $\cdot$ Self-Refine \\
\midrule
$\Delta$F1 & +0.0260 & +0.0194 & +0.0271 \\
licensed-add share & \textbf{107\%} & \textbf{99.5\%} & 0.7\% \\
unlicensed-add share & 4\% & $-$0.5\% & 17.0\% \\
unlicensed-delete share & $-$13\% & $-$2\% & 34.3\% \\
unresolved share & 1\% & 3\% & \textbf{48.3\%} \\
\bottomrule
\end{tabular}}
\end{table}

\paragraph{The unresolved component, reported proactively.}
For the eight settings with $|$contribution$| > 0.002$, unresolved items
drop from round 0 to final as follows: DocRED Self-Refine 1{,}629
$\rightarrow$ 575 (+0.0131, 48.3\% of its gain), Reflexion 1{,}459
$\rightarrow$ 658 (+0.0098, 37.3\%), CoVe 1{,}457 $\rightarrow$ 1{,}105
(+0.0039); Re-DocRED Self-Refine 1{,}308 $\rightarrow$ 341 (+0.0057),
Reflexion 1{,}152 $\rightarrow$ 457 (+0.0041); DWIE Reflexion 36
$\rightarrow$ 0 (+0.0038), Self-Refine 33 $\rightarrow$ 2 (+0.0032),
CoVe 24 $\rightarrow$ 3 (+0.0021). The second pass writes entity names
that match the documents' surface forms more often---a formatting
effect, not better extraction. Constraint loops are free of this
confound (unresolved component at most +0.0015).

\paragraph{Counting caliber.}
This appendix counts net $P_0 \rightarrow P_f$ edits; the
hit rates of feedback-named additions in the main text count at
round boundaries and include round-trip edits. The two calibers differ
by at most 0.014 in hit rate with identical conclusions; neither is
rewritten to match the other. Under the net caliber the DocRED and
Re-DocRED denominators coincide exactly (sr 359/359, cot+sr 132/132).

\section{Training and Implementation Details}
\label{app:training}

\subsection{Version Ledger}
torch 2.11.0+cu128, transformers 5.16.1, trl 1.12.0, peft 0.20.0,
Python 3.12. Base model \texttt{Qwen/Qwen3-8B}, revision
\texttt{b968826d9c46dd6066d109eabc6255188de91218}. Track A local
inference uses vLLM 0.11.2 in a separate environment (vLLM 0.30
conflicts with the driver; CUDA-13 builds break trl's GRPO import, so
the training environment has no vLLM). Track B evaluation runs under
transformers; the two sampling stacks differ and are recorded in every
run's \texttt{run\_config.json} together with the actually effective
and the silently ignored parameters.

\subsection{SFT Configuration}
Three variants (full / no-cons-field / gold-only) share one
configuration: 1{,}189 training samples (the intersection of the three
variant-specific sets, so that variants differ in fields, not data),
max\_seq\_len 5120 (DocRED) / 8192 (DWIE; longest retained document
7{,}830 tokens), LoRA $r{=}16$, $\alpha{=}32$, dropout 0.05, 447
steps over 3 epochs, completion-only loss via trl's prompt-completion
dataset (\texttt{completion\_only\_loss=True}; training aborts if the
flag is ignored). Each variant's prompt requests exactly the fields its
target contains---a four-field prompt with a two-field target teaches
the student to disobey the instruction (this bug once measured
gold-only's parse rate as 0\%). trl 1.12 silently ignores
\texttt{warmup\_ratio}; we therefore make no warmup claim. DWIE
training excludes 4 of 602 documents (3 overlength---10{,}113 / 8{,}723
/ 8{,}257 tokens---and 1 with an incomplete teacher output), leaving
598. On DocRED, GRPO runs on a fixed 500-document subset; on DWIE, SFT and
GRPO share the same 598 documents---a deliberate concession, since
splitting the small training set would underfeed both stages. 

\subsection{GRPO Configuration}
\label{app:grpo}
trl \texttt{GRPOTrainer} (the clipped token-level objective is the
library's; we describe configuration, not equations). Composite reward
$(\lambda_F, \lambda_C, \lambda_M) = (0.7, 0.2, 0.1)$ (extraction / consistency / format; verified against \texttt{run\_config}).  The weights were not swept under the current protocol, and we make no claim that they are optimal. Two observations bear on the consistency term: on DWIE, GRPO does not raise Cons (0.909 $\rightarrow$ 0.900), which suggests that its F1 gain there does not come through that term; on DocRED, Cons rises with F1 (0.567 $\rightarrow$ 0.607--0.667). Completions that trigger no rule neutralize the Cons term
(Appendix~\ref{app:eval}). Rollouts: temperature 0.8, cap 1024 tokens (8\% of SFT targets exceed
it; about 5.9\% of rollouts fail to parse and are reported as such),
$rp{=}1.1$;
evaluation: cap 3072, $rp{=}1.05$. Two learning rates, both reported:
$1{\times}10^{-5}$ (v6, cosine to $8{\times}10^{-8}$, KL 0.0007,
$\Delta$F1 $-0.0013$, $p{=}0.81$---ineffective) and $5{\times}10^{-5}$
(effective). Known unpatched reward loophole, with no observed impact:
an all-self-loop output scores $R{=}0.125$ above empty output's 0,
because the format term checks label validity but not resolver
acceptance; rejection rates did not rise during probes, and the term
will be scored on parsed triples if GRPO is extended.

\subsection{Teacher Traces}
1{,}200 DocRED documents (\${}24) and 602 DWIE documents (\${}7); the
DWIE traces were regenerated locally after the 9/19 loss (parse and
echo both 100\%). Echo fidelity: 99.9\% exact on DocRED, 100\% on DWIE
(residual deviations recorded in \texttt{teacher\_echo\_exact}). 11
unparseable teacher outputs (0.9\%) are excluded via the intersection;
11 gold-relationless documents are kept in the training set, as in
DocRED itself.

\subsection{The 9/19 Data Loss and \texttt{b4\_legacy}}
On 2026-09-19 the training server became unreachable, taking: the
DWIE teacher traces (regenerated locally), the SFT jsonl (rebuilt from
traces), 40 Track B run records with per-document caches and
predictions, and six LoRA adapters. DocRED-side Track B numbers in
Table~\ref{tab:trackb} are transcribed from experiment logs preserved
in \texttt{b4\_legacy.json} (footnoted in the table); DWIE-side records
are complete. The \texttt{b1\_sft.py} completion-only-loss patch was
lost with the server and has been reconstructed
(\texttt{patch\_b1\_completion\_only.py}); the DocRED Track B
checkpoints were produced by the lost original, so the reproduction
package documents the training script as a post-hoc reconstruction.

\subsection{Compute}
SFT $\approx$1.0 GPU-hour per variant (gold-only 0.82\,h); one GRPO
round (125 steps) $\approx$6.4\,h; the A10 grid totals $\approx$9.5
A100-80GB hours. Seeding the B8 ablation to three seeds would cost
$\approx$38 GPU-hours and cannot resolve its effect (trigger volume,
Table~\ref{tab:b8}), so it was not run. Table~\ref{tab:cost} lists costs
with the closure-addition ledger; teacher-trace generation (\${}24 for
DocRED, \${}7 for DWIE) is amortized into any break-even comparison with
API inference. Deterministic post-processing scores higher than iterative reflection on
DWIE in all three seeds, at $3.6\times$ lower cost (Section~\ref{sec:results:dwie}).

\begin{table}[t]
\caption{Cost and closure-addition ledger per configuration: closure
additions hitting/missing gold (seed 20261012), and cost per run (USD for
API runs, mean of three seeds where available, computed from logged token
counts at \$1.25/\$10 per million input/output tokens; GPU-hours for
local runs).}
\label{tab:cost}
\centering\small
\resizebox{\columnwidth}{!}{%
\begin{tabular}{lccc}
\toprule
Config & add-hit & add-miss & cost \\
\midrule
GPT-5.2 base+post (DWIE) & 254 & 90 & \$2.16 \\
GPT-5.2 cot+sr (DWIE) & 0 & 5 & \$7.85 \\
Qwen3-8B A10 grid (6 cells) & 268 & 2153 & 9.5 GPU-h \\
\bottomrule
\end{tabular}}
\end{table}

\section{Case Study: What the Feedback Looks Like}
\label{app:case}

We give one example as an illustration of the enumerator's feedback format, not as before–after evidence.

\paragraph{Example (DocRED, Monson, Maine).}
The document describes a four-level geographic hierarchy (Monson
$\subset$ Piscataquis County $\subset$ Maine $\subset$ United States).
A first-pass extraction produces only the locally adjacent containment
triples. The enumerator instantiates rule bodies over the predicted set
and reports, per rule, which instantiation fails to close and which
head fact is missing---e.g., for a containment chain, the two missing
skip-level facts. The reflection prompt renders exactly this list, and
the model's second pass can add the named facts.

\section{Comparison with Supervised DocRE Systems}
\label{app:sota}

Table~\ref{tab:sota} places our numbers next to supervised DocRE
systems, with each row's supervision inventory stated explicitly. The
comparison is \emph{not} head-to-head and we make no comparability
claim: supervised systems train on the benchmark's full training split
under a fixed label space and score given entity pairs, while our
systems extract generatively, and our protocol counts unresolved output
as false positives (Appendix~\ref{app:eval}), which theirs does not.
Supervised numbers are dev micro-F1 as reported in the original papers; our rows use the present
protocol (998 documents, micro-F1).

\begin{table}[t]
\caption{Supervision inventory and reported F1 on DocRED. sup.:
supervision source. DS: distant-supervision pretraining. Our rows are
not comparable to the supervised rows (see text).}
\label{tab:sota}
\centering\small
\resizebox{\columnwidth}{!}{%
\begin{tabular}{llllc}
\toprule
system & paradigm & backbone & sup.\ & F1 \\
\midrule
DREEAM~\cite{ma2023dreeam} & supervised PLM & RoBERTa-large & full train & 0.674 (dev) \\
KD-RoBERTa~\cite{tan-etal-2022-document} & supervised PLM + KD & RoBERTa-large & full train + DS & 0.671 (dev) \\
LMRC~\cite{li2024llmrelationclassifierdocumentlevel} & classifier + LoRA LLM & LLaMA3.1-8B & full train (cls + LoRA) & 0.611 (dev) \\
LoRA-only LLM~\cite{li2024llmrelationclassifierdocumentlevel} & pure LLM fine-tune & LLaMA2-13B & full train (LoRA) & 0.393 (dev) \\
GPT-4 0-shot~\cite{zhu2024refining} & prompting & GPT-4 & none & 0.156 (test) \\
GPT-4 multi-dim~\cite{zhu2024refining} & prompting & GPT-4 & none & 0.156--0.213 \\
\midrule
\textsc{Consistre} Track A & 5-shot prompting & GPT-5.2 & 5 demos & .2976 / .2741$^{*}$ (dev) \\
\textsc{Consistre} Track B & KD + RL alignment & Qwen3-8B & 1.2K SFT + 0.5K RL & .4313 (dev) \\
\bottomrule
\end{tabular}}
\end{table}

Two observations survive the protocol gap. First, the supervised upper
bound sits far above every LLM-based row, ours included---on a
near-closed reading of the benchmark, end-to-end LLM extraction is not
yet competitive with fixed-label classification, which is one more
reason to evaluate LLM-based methods by what they uniquely offer
(schema-free deployment and structural control) rather than by F1
alone. Second, supervision form matters more than model size: LMRC's
advantage over LoRA-only fine-tuning (0.611 vs.\ 0.393) comes from its
separately trained relation-candidate classifier, and our Track B row
uses roughly one third of the training documents with no auxiliary
classifier.

\section{Additional Related Work and Positioning}
\label{app:related}

\paragraph{Distillation and RL alignment.}
Our training-time track builds on knowledge
distillation~\cite{hinton2015distillingknowledgeneuralnetwork}, in particular rationale-style
distillation with reduced data~\cite{hsieh2023distillingstepbystepoutperforminglarger,gu2024minillm},
and on RL alignment: reinforcement learning from human or AI
feedback~\cite{ouyang2022traininglanguagemodelsfollow,bai2022constitutional}, optimized via
PPO~\cite{schulman2017proximal}, DPO~\cite{rafailov2024directpreferenceoptimizationlanguage}, or group-relative
policy optimization~\cite{shao2024deepseekmathpushinglimitsmathematical}, with process-level
supervision improving step validity~\cite{lightman2023let}.

\paragraph{Positioning.}
Three contrasts delimit our contribution. First, \emph{what the consistency
signal is}: BA-MCTS and ToolRE treat consistency as a scalar that an LLM
assigns to a triple set, and LogicReward's prover, even after its full
refinement pipeline, verifies only 50.8\% of reasoning steps and falls back
to token likelihood for the rest. Our check is deterministic rule matching
over the triple set with full coverage, and its feedback names which rule
fails to close and which fact is missing. Second, \emph{what rules are
for}: JMRL, RuAG, and SiftingLogic~\cite{heng-etal-2026-re2} use rules to improve or expand
annotations, and therefore need the rules to be true; we use rules as a
measurement instrument and only need them to be reproducible, with
admission backed by dual-split satisfaction rates. Notably, the schema that
SiftingLogic~\cite{heng-etal-2026-re2} uses to expand annotations includes directions our screen
rejects: \texttt{P17}${\Rightarrow}$\texttt{P131} is satisfied by only
12.8\% of body instantiations on DocRED gold, and
\texttt{P54}${\Rightarrow}$\texttt{P463} by 0.0\%. Third, \emph{what is
measured}: REPAIR and LOCO-LMs measure or improve the consistency of the
model itself; we measure the consistency of the \emph{gold annotations},
and quantify how the measured defect shrinks as annotations are revised.

\paragraph{Where rules should come from.}
Schema-derived rules screened by satisfaction rates are a fourth
provenance path, distinct from hand-crafted rule sets, rules mined from
annotations~\cite{qi-etal-2024-end,zhang2025ruag}, and rules generated at
runtime. Mining inherits annotation conventions: on DocRED gold,
\texttt{P150}${\Rightarrow}$\texttt{P131} is satisfied by 81.6\% of body
instantiations while the reverse direction reaches only 40.3\%---an
asymmetry of annotation practice, not of geography, that a miner would
read as rule confidence. Deriving rules from the schema instead asks
nothing of the annotations: the rule set is a function of the relation
inventory and a snapshot date, admission evidence is published per rule
(Appendix~\ref{app:rules}), and the rules are required to be
reproducible instruments, not truths.

\section{Reproduction Package}
\label{app:repro}

The package releases: (i) code---\texttt{rule\_engine.py} (the single
constraint enumerator shared by both tracks), \texttt{resolve.py},
\texttt{postproc.py}, \texttt{s1\_eval.py}, \texttt{audit\_schema.py},
\texttt{run\_schema.py}, \texttt{trace\_format.py}, the Track A drivers
(\texttt{a\_run\_docred.py}, \texttt{a2\_baselines.py},
\texttt{a\_analyze.py}), the Track B chain (\texttt{b0\_teacher\_traces.py},
\texttt{b\_dataset.py}, \texttt{b1\_sft.py},
\texttt{patch\_b1\_completion\_only.py}, \texttt{b2\_grpo.py},
\texttt{merge\_adapter.py}), and the two table generators
(\texttt{make\_tables.py}, \texttt{make\_run\_tables.py}); (ii) rule
files---\texttt{docred\_rules\_v1\_1.json} and
\texttt{dwie\_rules\_v1.1.json}, each pinned by sha256 and containing
per-rule grounding evidence (Appendix~\ref{app:rules}); (iii) document
ID files---\texttt{doc\_ids\_sft.txt}, \texttt{doc\_ids\_grpo.txt},
\texttt{s5\_splits.json}, and \texttt{\_doc\_ids.json} with
\texttt{demo\_ids}, all fixed before any model call under seed
20261012; (iv) \texttt{b4\_legacy.json}, the recovered Track B run records with its provenance note; (v)
analysis artifacts---\texttt{analysis\_w4.json},
\texttt{a12\_tests.json}, and the three A15
decomposition files. Wikidata snapshot date 2026-08-26; main seed
20261012, additional seeds 7/2024 (Track A) and 42/7/2024 (Track B).

\end{document}